\documentclass[11pt]{article}

\usepackage{acl}

\usepackage{times}
\usepackage{latexsym}
\usepackage[T1]{fontenc}
\usepackage{inconsolata}
\usepackage[utf8]{inputenc}
\usepackage{microtype}
\usepackage{amsmath}
\usepackage[table]{xcolor}

\usepackage{inconsolata}

\usepackage{graphicx}
\usepackage{multirow}
\usepackage{booktabs}
\usepackage[normalem]{ulem}
\usepackage{caption}
\usepackage{subcaption}
\usepackage{tikz}
\usepackage{float}
\usepackage{url}
\usepackage{tabularx}
\usepackage{cleveref}
\usepackage{hyperref}
\usepackage[inline]{enumitem}
\usepackage{soul}         % In your preamble
\sethlcolor{blue!20}

\title{Calibrating Beyond English: Language Diversity for Better Quantized Multilingual LLMs}

\author{Everlyn Asiko Chimoto$^{1}$, Mostafa Elhoushi$^{2}$, Bruce Bassett$^{3,4,5}$ \medskip \\
$^1$Lelapa AI, $^2$Cerebras Systems, Inc ,  $^3$University of the Witwatersrand,\\ $^{4}$University of Cape Town, South Africa, $^{5}$South African Astronomical Observatory \\ Correspondence: \href{mailto:email@domain}{everlyn.asiko@lelapa.ai} }

\begin{document}
\maketitle
\begin{abstract}
Quantization is an effective technique for reducing the storage footprint and computational costs of Large Language Models (LLMs), but it often results in performance degradation. Existing post-training quantization methods typically use small, English-only calibration sets; however, their impact on multilingual models remains underexplored. We systematically evaluate eight calibration settings (five single-language and three multilingual mixes) on two quantizers (GPTQ, AWQ) on data from 10 languages. Our findings reveal a consistent trend: non-English and multilingual calibration sets significantly improve perplexity compared to English-only baselines. Specifically, we observe notable average perplexity gains across both quantizers on Llama3.1 8B and Qwen2.5 7B, with multilingual mixes achieving the largest overall reductions of up to 3.52 points in perplexity. Furthermore, our analysis indicates that tailoring calibration sets to the evaluation language yields the largest improvements for individual languages, underscoring the importance of linguistic alignment. We also identify specific failure cases where certain language-quantizer combinations degrade performance, which we trace to differences in activation range distributions across languages. These results highlight that static one-size-fits-all calibration is suboptimal and that tailoring calibration data, both in language and diversity, plays a crucial role in robustly quantizing multilingual LLMs.
\end{abstract}

\section{Introduction}

Quantization -- where the numerical precision of model parameters is reduced (e.g., from 32-bit floats to 8-bit or lower) -- is a highly effective model compression technique and has become the de facto method for using Large Language Models (LLMs) on smaller, more accessible, and affordable infrastructure~\cite{720541,DBLP:journals/corr/abs-2103-13630,zhou2024surveyefficientinferencelarge}. However, its effectiveness depends critically on the calibration set used~\cite{pmlr-v139-hubara21a}. Current practice overwhelmingly relies on English-only calibration sets such as C4~\citep{10.5555/3455716.3455856} or Pile~\citep{pile} ~\cite{frantar2023gptqaccurateposttrainingquantization,lin2023awq}, even for multilingual models. This raises a key concern: does calibrating on English alone limit performance across other languages, and can we improve multilingual performance by broadening the choice of calibration set?

\begin{figure}
    \centering
    \resizebox{\columnwidth}{!}{
        \includegraphics{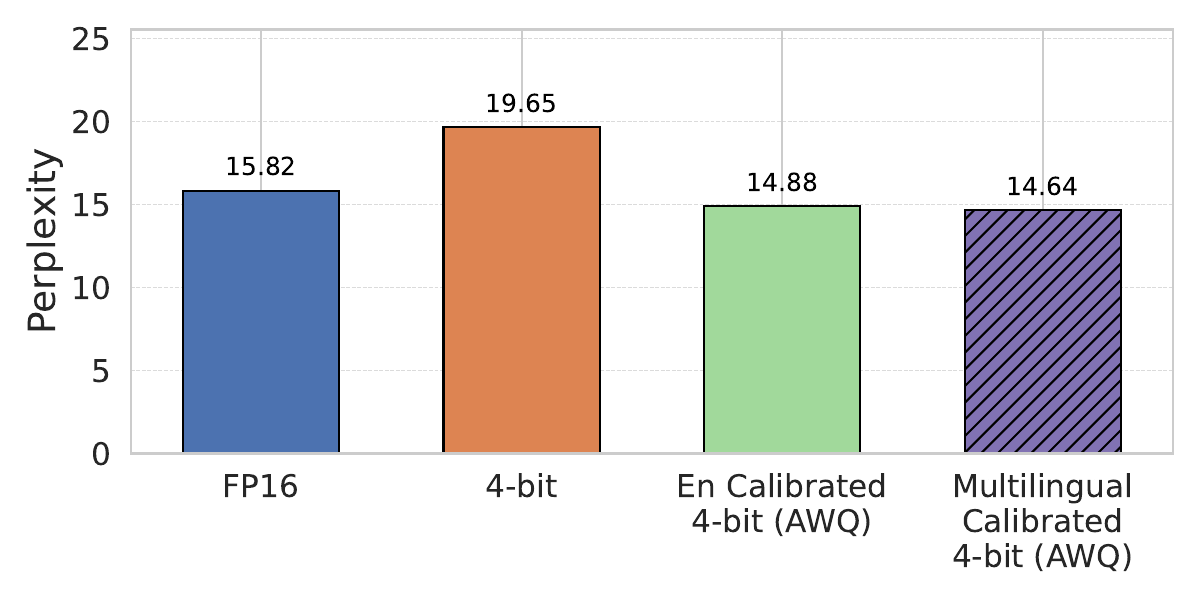}
        }
        \caption{Average perplexity on 10 languages for Llama3.1 8B. Multilingual calibration achieves the lowest perplexity (14.64), illustrating that calibration language affects quantization quality.}
        \label{fig:setup}
\end{figure}

This question matters most in settings where compute is scarce. In particular, techniques like quantization are often necessary for deploying models in low-resource language settings \cite{ahia-etal-2021-low-resource}. Sadly, prior work has shown that quantization disproportionately degrades performance in multilingual LLMs, with the most severe effects observed for low-resource, and non-Latin script languages -- and as such, it provides limited utility for the long tail of low-resource languages~\cite{marchisio-etal-2024-quantization}.

Figure~\ref{fig:setup} provides a motivating example. It shows the perplexity score for Llama3.1 8B under 4-bit quantization with AWQ. While weight-only 4-bit quantization degrades performance relative to FP16, using a calibration set makes a clear difference: English-only calibration reduces perplexity to 14.88, but a multilingual calibration set lowers it further to 14.64. The improvement is modest in absolute terms, yet consistent with a broader pattern across languages: calibration language and composition matter.

\begin{figure*}[htbp!]
  \centering

  \begin{subfigure}[b]{0.45\textwidth}
    \includegraphics[width=\linewidth]{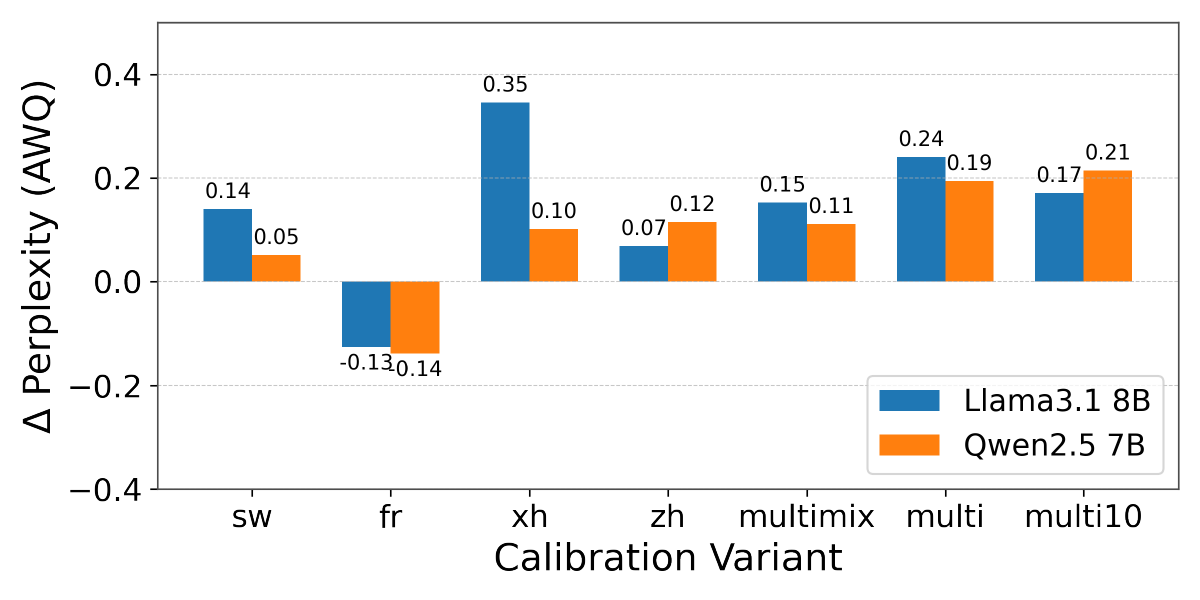}
    \caption{AWQ Delta}
    \label{fig:awq_de}
  \end{subfigure}
  \begin{subfigure}[b]{0.45\textwidth}
    \includegraphics[width=\linewidth]{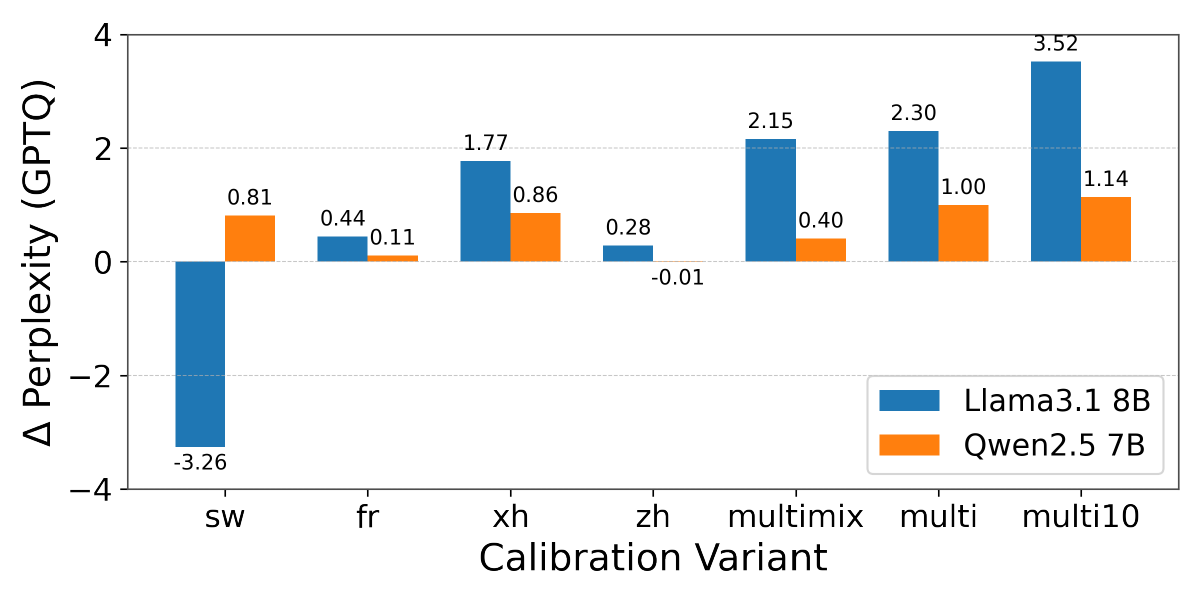}
    \caption{GPTQ Delta}
    \label{fig:gptq_sw}
  \end{subfigure}
  \caption{$\Delta$ Perplexity (higher = better) on the Wikipedia multilingual test set (English, French, Swahili, Xhosa, Chinese, Sesotho, Yoruba, Zulu, Hausa, Igbo), relative to an English-only calibration baseline. Calibration with non-English languages lead to better perplexity than English-only calibration. Multilingual variants(refer to \Cref{sec:data} for details) provide the largest gains, showing that a linguistically diverse calibration set can outperform an English-only baseline.}
  \label{fig:Delta}
\end{figure*}

We hypothesize that non-English and mixed-language calibration sets can better preserve performance across diverse languages and might be an avenue for improvement for low-resourced languages. To test this, we compare English, non-English (French, Swahili, Chinese, isiXhosa), and multilingual calibration sets in post-training quantization. We evaluate these calibration sets on two similar sized LLMs (Llama3.1 8B, Qwen2.5 7B) using two Post-Training Quantization (PTQ) methods: (GPTQ, AWQ). Beyond performance, we probe the models’ internal behavior (quantization error, weight updates, and activation distributions) when the calibration set language is varied. Our analysis also asks what constitutes an optimal calibration set in terms of token distribution and linguistic coverage.

We organize our investigation around three key arguments, each tied to a research question denoted as RQ:

\begin{itemize}
  \item \textbf{Argument 1:} “One-Size-Fits-All” Calibration is Suboptimal
  
    \textbf{RQ1:} How does calibration-set language/composition affect quantization accuracy across languages?  
    
    \textbf{Result:} We show that non-English and multilingual sets can outperform English-only sets (up to +3.52 ppl. on Llama-GPTQ).
    %\end{itemize}

  \item \textbf{Argument 2:} Calibration Sets Must Account for Rare and Extreme Data
  
    \textbf{RQ2:} Do outlier tokens or extreme activations in calibration data drive quantization error?  
    
    \textbf{Result:} We demonstrate that including rare tokens and activation outliers reduces performance degradation.
    % \end{itemize}

  \item \textbf{Argument 3:} Calibration effectiveness is quantizer-dependent.
  
    \textbf{RQ3:} How do different calibration sets interact with GPTQ’s Hessian-based updates versus AWQ’s activation scaling?  

    \textbf{Result:} GPTQ is more sensitive to calibration-language shifts, while AWQ is more robust due to its rescaling design.
\end{itemize}

\section{Background:  Calibration Sets for Quantization}

Popular Post-Training Quantization (PTQ) techniques utilize calibration sets to determine how the weights are quantized. Examples include GPTQ, which changes weights in each channel and dynamically adjusts the quantized weights to compensate for the error, and AWQ which uses the calibration set to identify salient weights and scales them to keep their magnitude high. In this study, we focus our investigation on GPTQ~\cite{frantar2023gptqaccurateposttrainingquantization} and AWQ~\cite{lin2023awq} for their prominence and representation in weight-only PTQ as well as widespread adoption in popular LLM libraries (e.g., GPTQModel\footnote{https://github.com/ModelCloud/GPTQModel}, AutoGPTQ\footnote{https://github.com/AutoGPTQ/AutoGPTQ}, AutoAWQ\footnote{https://github.com/casper-hansen/AutoAWQ}, LLM-AWQ\footnote{https://github.com/mit-han-lab/llm-awq}). We discuss them below:

\begin{figure*}[htbp!]
  \centering

  \begin{subfigure}[b]{0.45\textwidth}
    \includegraphics[width=0.95\linewidth]{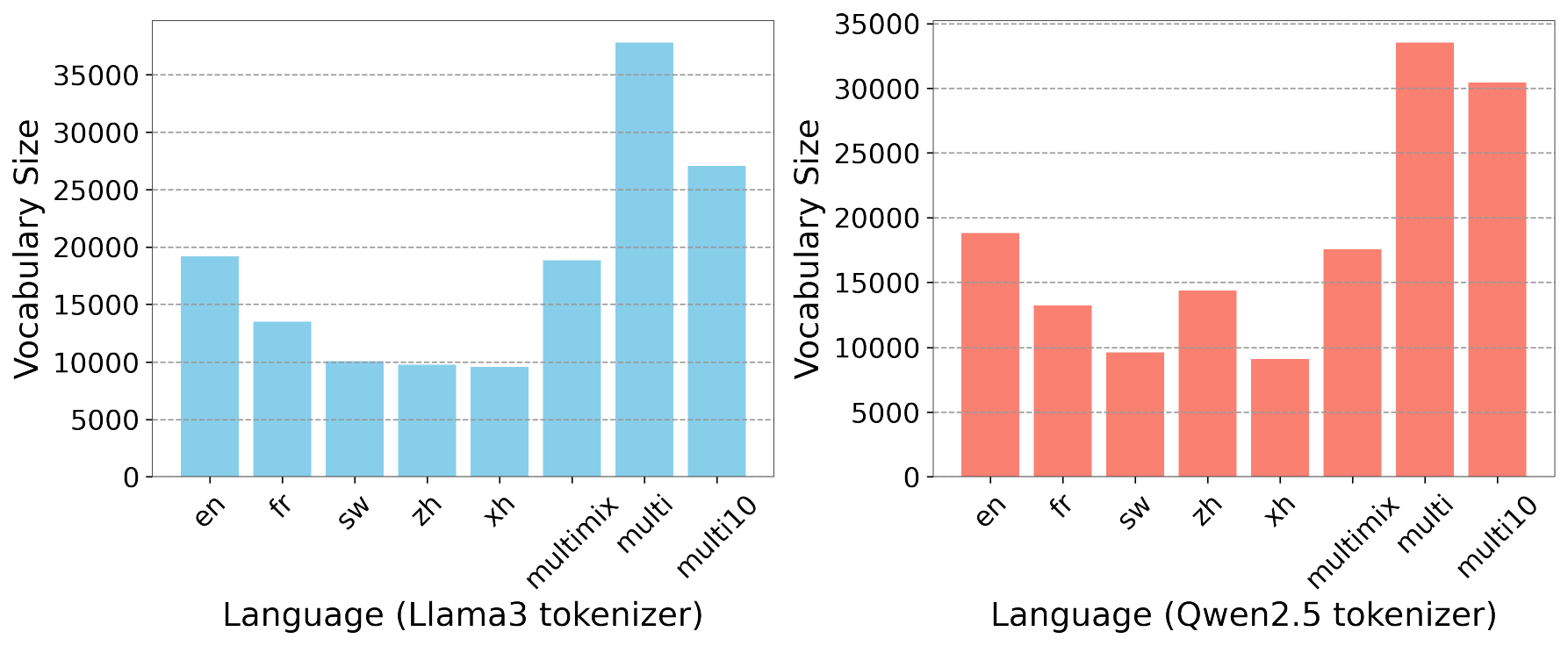}
    \caption{Unique vocabulary size per calibration set}
    \label{fig:vocab1}
  \end{subfigure}
  \begin{subfigure}[b]{0.45\textwidth}
    \includegraphics[width=0.95\linewidth]{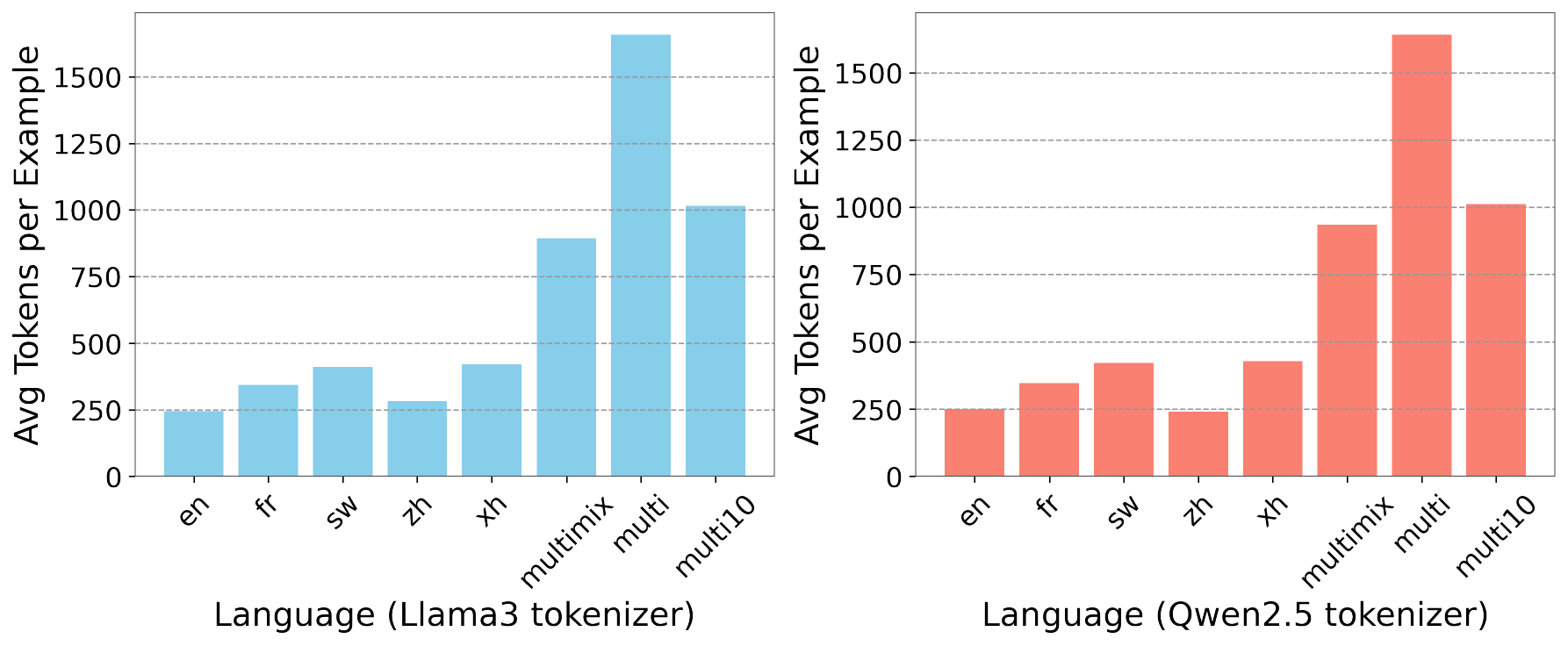}
    \caption{Mean number of tokens per calibration example}
    \label{fig:token}
  \end{subfigure}
  \caption{Comparison of calibration‐set statistics across different languages.  (a) The total unique tokens in each calibration set;  
  (b) The average token count per example. Multilingual sets exhibit both larger vocabularies and longer examples, indicating broader coverage of token contexts. This richer distribution correlates with the improved performance shown in \Cref{fig:Delta}.}
  \label{fig:vocab}
\end{figure*}

\paragraph{GPTQ} - General Post Training Quantization works in the following way: Given a weight matrix $W$ and calibration data $X$, we obtain the quantized weight $\hat{W}$ by computing the inverse Hessian via Cholesky decomposition defined by:
\[
H^{-1}
\;=\;
\bigl(2\,X X^{T} \;+\;\lambda I_{d}\bigr)^{-1},
\]
where \(\lambda>0\) is a damping factor and \(I_{d}\) is the \(d\times d\) identity matrix. 
We partition the columns of \(W\) into non‐overlapping groups of size \(B\). GPTQ then maps each row’s real-valued weights into a small fixed set of integer levels defined by  asymmetric min–max. This is referred to as the quantization grid. For each group, beginning at column index \(i\), we quantize its columns \(j = i, i+1,\dots,i+B-1\) in sequence, by rounding each to the nearest number in the selected quantization grid, denoted by \(Q[:,j]\). The rest of the weights are adjusted using error, $E$: 
  \[
    E[:,j-i]
    \;=\;
    \frac{W[:,j] - Q[:,j]}{H^{-1}[j,j]},
  \]
where $i$ is the current column index, \(H^{-1}[j,j]\) is the \([j,j]\) entry of \(H^{-1}\). The updated weights are:
$$W[:,j:(i+B)] -= E[:,j-i] * H^{-1}[j,j:(i+B)].$$

This continues until all weights in each group have been updated. This procedure incrementally minimizes the quantization error by using the Hessian approximation to propagate residuals within each group.

\paragraph{AWQ} - Activation Aware Quantization works by identifying salient channels based on activation magnitudes. It runs a calibration set through the model to collect per-input-channel activation magnitudes. The channels with the largest average activations are deemed most salient. For each salient channel \(j\in\mathcal{S}\), we scale its corresponding weights in \(W[:,j]\) by a factor \(\alpha_j>1\) (e.g., proportional to \(\max |W[:,j]|\)), yielding a scaled weight \(w'_{r,j}\).  AWQ then applies standard grouped PTQ (e.g., 3- or 4-bit) on each row or group of size \(g\).
This two‐step process, salient‐channel scaling followed by grouped grid quantization, preserves the most influential weights while compressing the remainder.

\begin{figure*}[htbp!]
  \centering

  \begin{subfigure}[b]{0.49\textwidth}
    \includegraphics[width=1.1\linewidth]{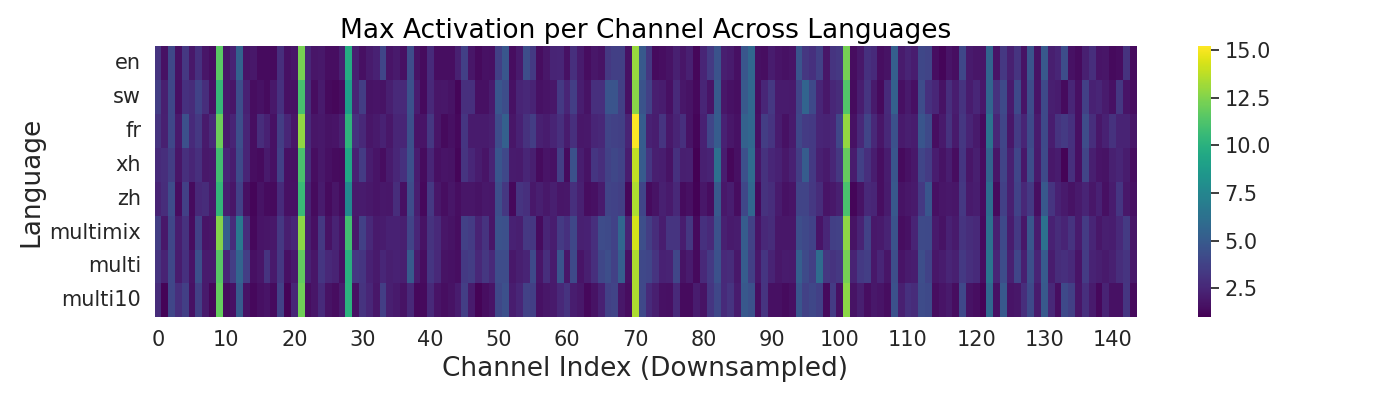}
    \caption{Gate projection max channel activations}
    \label{fig:gate_project}
  \end{subfigure}
  \begin{subfigure}[b]{0.49\textwidth}
    \includegraphics[width=1.1\linewidth]{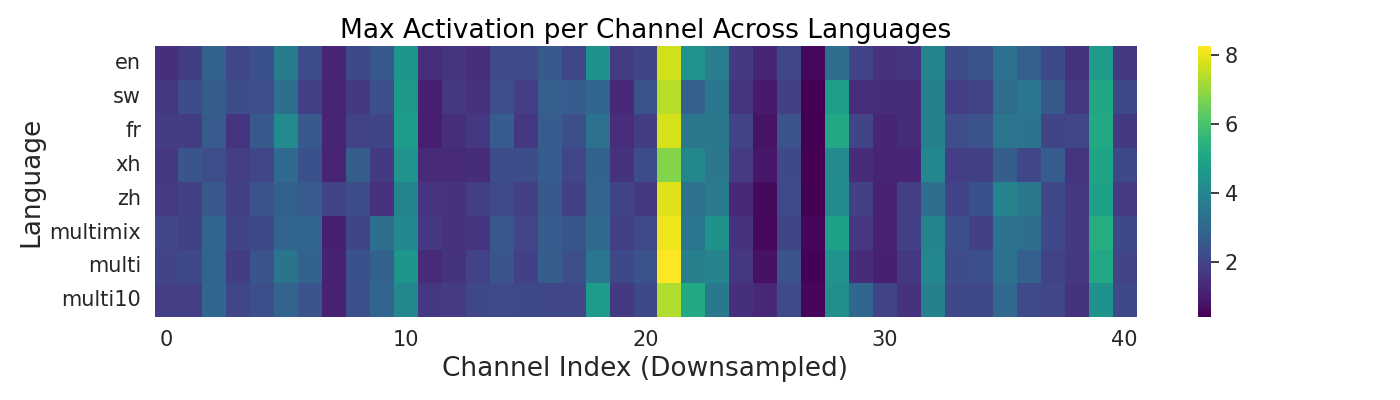}
    \caption{Query projection max channel activations}
    \label{fig:q_project}
  \end{subfigure}
  \caption{Heatmaps of maximum activations after AWQ quantization of Llama3.1 8B for (a) the MLP gate projection and (b) the attention query projection in layer 31 (selected for its large quantization errors; see \Cref{sec:appendix}). Rows denote different calibration language variants. Across languages, the same salient channels dominate, but their peak magnitudes shift—showing that AWQ rescales fixed channels rather than changing which channels matter, which helps explain the modest perplexity deltas. }
  \label{fig:channel}
\end{figure*}

\paragraph{Comparison Studies} - While a growing body of work has examined quantization for LLMs, only a few have systematically investigated the role of calibration data in shaping post-training quantization outcomes. \citet{williams-aletras-2024-impact} conducted a large-scale study of calibration data, showing that performance varies markedly across calibration sources and even across sets from the same source. Unlike their focus on English-only web data, we investigate multilingual calibration sets and their role across quantizers. \citet{marchisio-etal-2024-quantization} analyzes multilingual models, finding that quantization disproportionately harms non-Latin script languages, that automatic metrics underestimate degradation, and that reasoning tasks degrade fastest. While they highlight disparities, we explicitly compare multilingual versus monolingual calibration sets to measure how calibration choice influences these effects. \citet{williams-etal-2025-self} propose self-calibration, using the model itself to generate synthetic calibration sets, which often match or surpass real data. \citet{pmlr-v267-elhoushi25a} showed that calibrating using single diverse curated sample could outperform a dataset. Our work differs by isolating language composition as the key experimental variable in quantization calibration. While keeping everything else constant—the model, quantizer, calibration size, and domain—we vary only the languages used for calibration. To our knowledge, this is the first systematic study showing that multilingual calibration is a primary control factor for quantization stability, not just a secondary consideration. We test this across different quantizers and model families. Beyond measuring accuracy, we analyze quantization error, activation distributions, and Hessian statistics to explain why language-aware calibration improves robustness, especially for low-resource languages.

\section{Experimental Set-up}

\subsection{Models}

We assess the impact of multilingual calibration sets using two pre-trained multilingual LLMs: Llama3.1 8B (instruction-tuned)\footnote{HuggingFace ID: {\tt meta-Llama/Llama3.1 8B-instruct}} and Qwen2.5 7B (instruction-tuned)\footnote{HuggingFace ID: {\tt qwen/Qwen-2.5-7B-Instruct}}. They are 8B and 7B-parameter respectively. We selected these models for their multilingual properties, each being trained on 8~\cite{grattafiori2024Llama3herdmodels,Llama3_8b_instruct} and 29~\cite{qwen2025qwen25technicalreport,qwen2_5_blog} languages respectively. To verify that the observed calibration-language effects are not specific to these model families, we additionally include BLOOMZ-7B1-MT\footnote{HuggingFace ID: {\tt bigscience/bloomz-7b1-mt}} as a multilingual-heavy baseline, trained on 46 languages~\cite{muennighoff-etal-2023-crosslingual}. Due to space constraints, BLOOMZ-7B1-MT results are reported in the appendix (\Cref{sec:appendix_bloom}). All experiments were performed on a single NVIDIA A100 40GB GPU. We utilised the official pre-trained weights available on the Hugging Face Hub for both models.

\subsection{Quantization Technique}

We primarily apply two post-training weight quantization methods in our experiments: GPTQ and AWQ (as discussed in the Background). To assess whether calibration-language effects generalize beyond these widely used approaches, we additionally evaluate Any4, a recent calibration-based 4-bit quantization method~\cite{pmlr-v267-elhoushi25a}; detailed Any4 results are reported in \Cref{sec:any4}. For GPTQ, we follow an existing open-source implementation \citet{qubitium2024gptqmodel} adapted from \cite{frantar2023gptqaccurateposttrainingquantization} which uses 4-bit quantization and group size 128. We set the calibration batch size to 2 in our GPTQ procedure. Similarly, for AWQ, we use 4-bit quantization and group size 128 \citep{lin2023awq}. Prior work has shown that 4-bit quantization offers a favorable balance between model size and performance~\citep{pmlr-v202-dettmers23a}; thus, we fixed the bit width to 4 bits for all quantization methods in our experiments. 

\subsection{Calibration Data Sources and Preparation}
\label{sec:data}

We construct eight calibration sets for each language or strategy, designed to isolate the effect of \emph{calibration language composition} under a fixed calibration budget:
\begin{enumerate*}[label=(\roman*)]
  \item \emph{English baseline}: the original English calibration sets—C4 for GPTQ and the Pile validation split for AWQ. For AWQ, we use 512 examples of 512 tokens each; for GPTQ, we use 1024 examples of 1042 tokens each.
\end{enumerate*}

Across all experiments, we explicitly hold the total calibration token budget constant to decouple the effect of calibration language composition from calibration size. Prior work in post-training quantization demonstrates that, beyond a moderate budget, performance is substantially more sensitive to the \emph{distributional properties} of calibration data than to the absolute number of tokens (e.g., GPTQ, AWQ, SmoothQuant; \citealp{williams-aletras-2024-impact}). Accordingly, our goal is not to optimize calibration size, but to conduct controlled experiments that disentangle the impact of linguistic diversity under a fixed budget.

\begin{enumerate*}[label=(\roman*), resume]
  \item \emph{Single-language translated}: translations of the English calibration set into French, Swahili, Chinese, and isiXhosa using the \texttt{deep-translator} library\footnote{https://pypi.org/project/deep-translator}. These sets provide a controlled mechanism for isolating language effects while holding lexical content constant.
  
  \item \emph{C4}: randomly sampled examples from the C4 training split in each target language.
  
  \item \emph{Wikipedia}: randomly sampled articles from the Wikipedia training split, excluding the first 5{,}000 examples reserved for evaluation.
\end{enumerate*}

Unlike translated calibration sets, C4 and Wikipedia provide \emph{native multilingual data} with naturally occurring lexical, syntactic, and orthographic variation, allowing us to verify that observed effects are not artifacts of machine translation. Full results are reported in \Cref{sec:appendix_translation}.

\begin{enumerate*}[label=(\roman*), resume]
  \item \emph{C4 Multilingual, \texttt{multimix}}: random samples drawn from the C4 multilingual split without language balancing.
  
  \item \emph{C4 Multilingual, \texttt{multi10}}: an equal allocation of \(\frac{N}{10}\) examples from each of ten languages (\texttt{en}, \texttt{fr}, \texttt{sw}, \texttt{zh}, \texttt{xh}, \texttt{st}, \texttt{zu}, \texttt{yo}, \texttt{ig}, \texttt{ha}).
  
  \item \emph{C4 Multilingual, \texttt{multi}}: \(N\) examples drawn uniformly from all 112 C4 languages\footnote{https://huggingface.co/datasets/allenai/c4}.
  
  \item \emph{Code/Math variants (\texttt{code-*}, \texttt{math-*}, \texttt{codemath-*})}: to assess whether rare or structured tokens (e.g., numerals, symbols, identifiers) aid quantization, we augment each base calibration set \(X\) with uniformly sampled data from DeepMind’s \texttt{mathematics\_dataset}~\cite{2019arXiv} and CodeParrot’s \texttt{github-code-clean}\footnote{\url{https://huggingface.co/datasets/codeparrot/github-code-clean}}. All variants preserve the original calibration token budget via uniform mixing.
\end{enumerate*}

\subsection{Evaluation Datasets and Languages}

We evaluate quantization performance across multiple languages and evaluation datasets. The primary evaluation languages are English (\texttt{en}), French (\texttt{fr}), Swahili (\texttt{sw}), Chinese (\texttt{zh}), and isiXhosa (\texttt{xh}). In addition, we also evaluate 5 more low-resource languages: Sotho (\texttt{st}), Zulu (\texttt{zu}), Yoruba (\texttt{yo}), Igbo (\texttt{ig}), and Hausa (\texttt{ha}).

For each language, we measure performance on:
\begin{enumerate*}[label=(\roman*)]
  \item Wikipedia validation; first 5,000 samples of the available split,
  \item C4 validation set.  
\end{enumerate*} For each model/quantizer/calibration combination, we report:
\begin{itemize}[nosep]
  \item Absolute perplexity on FP16 (float-16 precision), 4-bit weight-only baseline, vs.\ quantized model. 
  
  FP16 represents a standard full-precision baseline, while “4-bit weight-only” indicates that all weights have been rounded to 4 bits but activations remain in FP16.  Comparing these shows the accuracy loss from quantization. 
  \item \(\Delta\) perplexity relative to the English-only calibration baseline.

  We focus on perplexity in the main text. Any downstream task probes are deferred to the appendix and do not affect our primary conclusions.
  \item Downstream task accuracy on XNLI \cite{conneau-etal-2018-xnli}, XStoryCloze \cite{lin-etal-2022-shot}, and Global MMLU \cite{singh-etal-2025-global}. XNLI covers sentence-level natural language inference in \texttt{en}, \texttt{fr}, \texttt{sw}, and \texttt{zh}; XStoryCloze evaluates paragraph-level causal reasoning in \texttt{en}, \texttt{sw}, and \texttt{zh}; and Global MMLU provides a multilingual knowledge and reasoning benchmark spanning diverse languages and domains. Due to computational constraints, downstream evaluations are reported for the languages supported in the experiments. 
  \item Layer-wise quantization error and activation statistics.  
\end{itemize}

\section{Results and Analysis}

\begin{table}[t]
\centering
\resizebox{\columnwidth}{!}{
\begin{tabular}{lccc}
\toprule
\textbf{Calibration} & \textbf{XNLI Avg (N=5)} & \textbf{XStoryCloze Avg (N=4)} & \textbf{GlobalMMLU Avg (N=5)} \\
\midrule
\textbf{en}       & 45.51 $\pm$ 0.45 & 64.97 $\pm$ 0.61 & 48.90 $\pm$ 1.07 \\
\textbf{sw}       & \textbf{46.19 $\pm$ 0.45} & 62.99 $\pm$ 0.62 & 44.30 $\pm$ 1.07 \\
\textbf{fr}       & 45.43 $\pm$ 0.45 & 65.36 $\pm$ 0.61 & 48.25 $\pm$ 1.08 \\
\textbf{xh}       & 45.18 $\pm$ 0.45 & 64.91 $\pm$ 0.61 & 47.15 $\pm$ 1.09 \\
\textbf{zh}       & 45.15 $\pm$ 0.45 & 65.48 $\pm$ 0.60 & 49.25 $\pm$ 1.07 \\
\textbf{multi}    & 45.54 $\pm$ 0.45 & 65.56 $\pm$ 0.61 & \textbf{50.45 $\pm$ 1.08} \\
\textbf{multi10} & 45.49 $\pm$ 0.45 & \textbf{65.82 $\pm$ 0.61} & 50.20 $\pm$ 1.09 \\
\textbf{multimix} & 44.96 $\pm$ 0.45 & 65.06 $\pm$ 0.61 & 49.95 $\pm$ 1.07 \\
\bottomrule
\end{tabular}}
\caption{Task-level performance (macro-averaged) using GPTQ quantization. Avg denotes an unweighted macro-average over languages available for each task (XNLI: 5, XStoryCloze: 4, MMLU: 5). Error bars are propagated from per-language standard errors assuming independence.}
\label{tab:gptq4bit_downstream_avg}
\end{table}

\begin{figure}
    \centering
    \resizebox{\columnwidth}{!}{
        \includegraphics{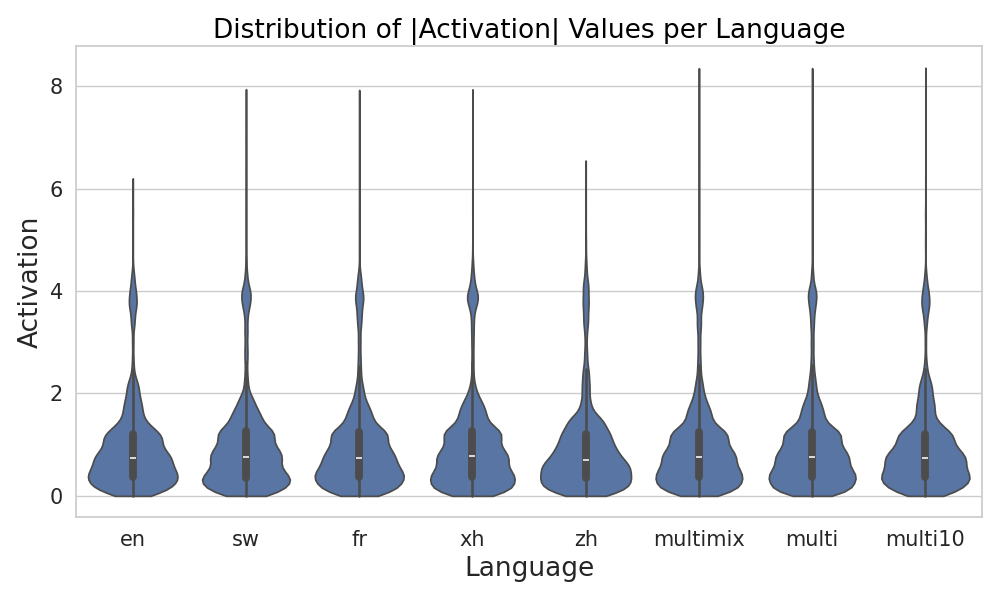}
        }
        \caption{Activation distributions from the unquantized Llama model across different calibration sets. Violin plots compare absolute activations per set: the three multilingual variants (\texttt{multimix}, \texttt{multi}, \texttt{multi10}) exhibit longer upper tails than single-language sets (\texttt{en}, \texttt{fr}, \texttt{sw}, \texttt{zh}, \texttt{xh}), capturing higher-magnitude outliers. This broader coverage suggests that multilingual calibration is better suited to handle unseen extremes at test time.}
        \label{fig:multiact}
\end{figure}

\hl{\textbf{Argument 1}: One-Size-Fits-All Calibration Set is Suboptimal.}

\textit{Using a non-English calibration set generally leads to better post-quantization performance than using an English-only calibration set.}

\begin{figure*}[htb!]
  \centering
  \begin{subfigure}[b]{0.45\textwidth}%{1\columnwidth}
    \includegraphics[width=\linewidth]{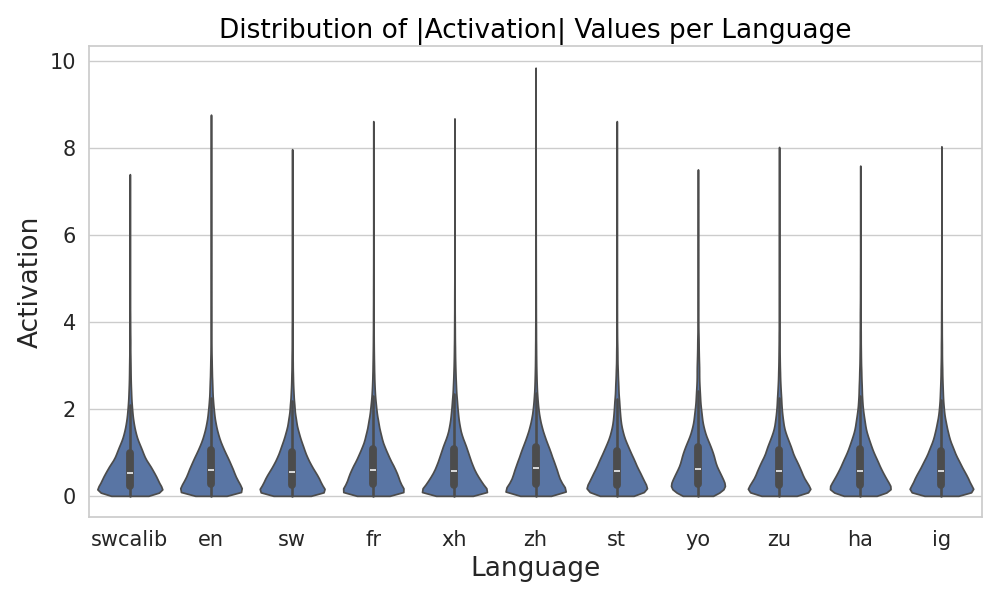}%sw2.pdf
    \caption{Activation distributions - Swahili calibration set (swcalib) v.s. Wikipedia test data.}
    \label{fig:sw1}
  \end{subfigure}
  \begin{subfigure}[b]{0.45\textwidth}%{1\columnwidth}
    \includegraphics[width=\linewidth]{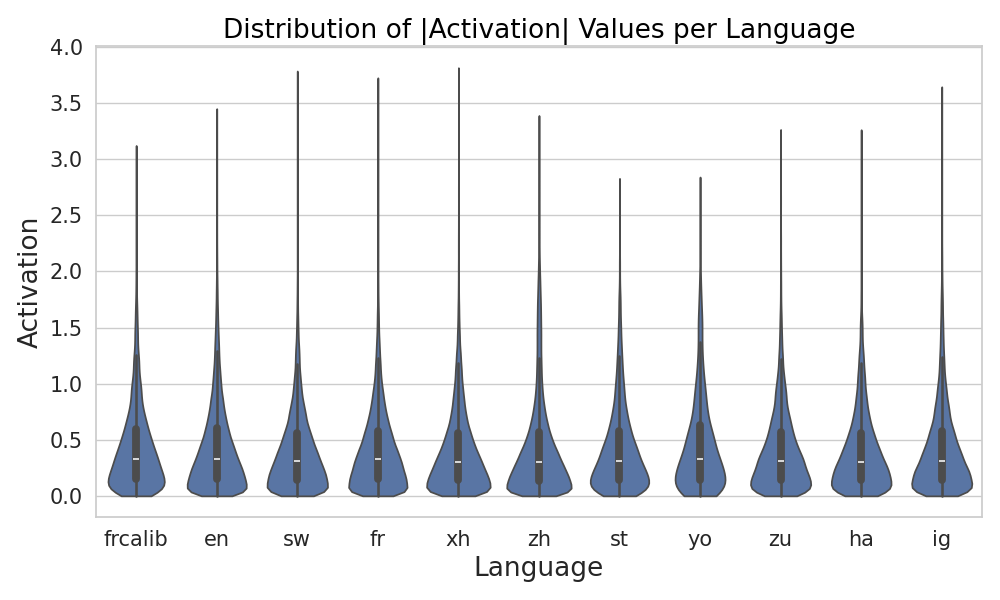}%fr_gptq.pdf
    \caption{Activation distributions - French calibration set (frcalib) vs. Wikipedia test data.}
    \label{fig:sw3}
\end{subfigure}
  \caption{Activation-range mismatch between calibration and test sets in the last Llama3.1 8B layer. (a) With Swahili calibration, activations span up to $\approx$7.5, while most test languages exceed 8. (b) With French calibration, activations top out at $\approx$3.2, while test languages extend beyond 3.3. Missing these outliers leads to overly conservative quantization thresholds, limiting perplexity gains on test languages with broader activation ranges.}
  \label{fig:swact}
\end{figure*}

In \Cref{fig:Delta}, we plot the change in perplexity: $ \Delta \mathrm{PPL} = \mathrm{PPL}_{\text{English}} - \mathrm{PPL}_{\text{Other}}$ on a 5000-sentence Wikipedia validation set for each language. Here, a positive delta means a reduction in perplexity (lower average loss per token), which corresponds to better model fit on the evaluation data. We see that almost all of the non-English calibration sets yield a positive $\Delta$ perplexity compared to the English calibration baseline. 
Notable exceptions are the French calibration set under AWQ and the Swahili calibration set under GPTQ on the Llama model, where performance decreases slightly relative to using the English set. We examine these two outlier cases in our analysis in \Cref{fig:swact} and see that Swahili and French calibration sets exhibit narrower ranges in their activation distribution compared to the broader distribution encountered during inference. This may result in poor representation of critical outliers as out-of-distribution values are clipped to the boundaries seen during calibration, resulting in substantial quantization error. Overall, the trend supports our hypothesis that English-only calibration is suboptimal for multilingual models.

\textit{Calibration Aligned with Evaluation Language Outperforms Static Sets.} Using a calibration set that matches the evaluation language yields significant gains on that language’s performance for Llama3.1 8B. As shown in \Cref{tab:Llama} and \Cref{tab:QWEN}, language‐specific calibration sets produce the largest perplexity reductions for their corresponding languages under AWQ. For instance, the Swahili calibration set yields an improvement on Swahili benchmarks, and similarly, the French, Chinese, and isiXhosa calibration sets achieve the highest gains for their respective languages. Moreover, calibration data can benefit closely related languages: the isiXhosa calibration set improves performance for both isiXhosa and its linguistic relatives Sotho and Zulu. This cross‐language uplift is most pronounced for Llama3.1 using AWQ, where isiXhosa calibration reduces perplexity by on Sotho and on Zulu (see Table~\ref{tab:Llama}). To verify that the observed calibration-language effects are not model-specific, we additionally conducted experiments on BLOOMZ-7B1-MT, a natively multilingual language model. While BLOOMZ-7B1-MT exhibits greater robustness to calibration-language choice compared to LLaMA, non-English calibration sets still consistently outperform English-only calibration, confirming the generality of our findings (see ~\Cref{sec:appendix_bloom}).

\textit{Perplexity gains from non-English calibration extend to downstream tasks.}
To assess whether perplexity improvements translate into functional gains, we evaluate GPTQ-quantized Llama3.1 8B on zero-shot XNLI, XStoryCloze, and Global MMLU (\Cref{tab:gptq4bit_downstream_avg}). Downstream performance broadly follows the perplexity trends observed above, albeit with smaller margins. On XNLI, calibration aligned with the evaluation language performs best, with Swahili calibration achieving the highest mean accuracy. In contrast, XStoryCloze favors calibration diversity: the \texttt{multi10} calibration set attains the highest average accuracy, outperforming English-only calibration. Global MMLU exhibits a similar pattern, with multilingual calibration consistently surpassing the English baseline. Across all benchmarks, both language-matched and multilingual calibration strategies outperform English-only calibration, indicating that calibration choice affects downstream task behavior rather than merely language modeling loss. We report a rank-correlation analysis between perplexity and downstream accuracy based on the XStoryCloze and Global MMLU results in \Cref{sec:perp_downcorrelation}. While limited to these two benchmarks, which offer consistent multilingual coverage and comparable evaluation structure, the analysis reveals a stable negative correlation, supporting the conclusion that calibration strategies which reduce perplexity also tend to yield higher downstream accuracy.

\begin{table*}[htbp!]
\centering
\resizebox{\textwidth}{!}{
\begin{tabular}{l|l|rrrrrrrrrrr}
\textbf{Quantization} & \textbf{Calibration} &
  \multicolumn{1}{l}{\textbf{en}} &
  \multicolumn{1}{l}{\textbf{sw}} &
  \multicolumn{1}{l}{\textbf{fr}} &
  \multicolumn{1}{l}{\textbf{xh}} &
  \multicolumn{1}{l}{\textbf{zh}} &
  \multicolumn{1}{l}{\textbf{st}} &
  \multicolumn{1}{l}{\textbf{yo}} &
  \multicolumn{1}{l}{\textbf{zu}} &
  \multicolumn{1}{l}{\textbf{ha}} &
  \multicolumn{1}{l}{\textbf{ig}} &
  \multicolumn{1}{l}{\textbf{Avg}} \\ \toprule
\textbf{FP16}            & --               & 7.327 & 6.510 & 5.698 & 69.300 & 9.526  & 17.579 & 10.148 & 13.410 & 11.013 & 7.645 & 15.816 \\
\textbf{Uniform INT4}    & --               & 8.327 & 7.836 & 6.323 & 84.210 & 11.085 & 23.266 & 12.551 & 17.053 & 15.670 & 10.151 & 19.647 \\ \hline
\textbf{AWQ}             & \textbf{en}      & 7.695 & 5.936 & 5.939 & 66.536 & 9.989  & 15.941 & 8.822  & 10.198 & 9.915  & 7.816 & 14.879 \\
\textbf{AWQ}             & \textbf{sw}      & \underline{7.679} & \cellcolor{purple!20}\textbf{5.859} & 5.944 & 65.762 & 9.983  & 15.702 & 8.804  & 10.103 & 9.859  & 7.695 & 14.739 \\
\textbf{AWQ}             & \textbf{fr}      & 7.693 & 5.950 & \cellcolor{purple!20}\textbf{5.913} & 67.328 & 9.975  & 16.127 & 8.841  & 10.365 & 9.997  & 7.855 & 15.004 \\
\textbf{AWQ}             & \textbf{xh}      & 7.703 & 5.914 & 5.951 & \cellcolor{purple!20}\textbf{64.254} & 10.005 & \cellcolor{yellow!20}\textbf{15.524} & \underline{8.721} & \cellcolor{purple!20}\textbf{9.811} & 9.815 & \cellcolor{purple!20}\textbf{7.630} & \cellcolor{purple!20}\textbf{14.533} \\
\textbf{AWQ}             & \textbf{zh}      & 7.683 & 5.924 & 5.938 & 66.004 & \cellcolor{purple!20}\textbf{9.888} & 15.901 & 8.887  & 10.126 & 9.982  & 7.772 & 14.810 \\
\textbf{AWQ}             & \textbf{multimix}& 7.683 & 5.913 & 5.929 & 65.662 & 9.970  & 15.702 & 8.789  & 10.050 & 9.851  & 7.724 & 14.727 \\
\textbf{AWQ}             & \textbf{multi}   & \textbf{7.675} & 5.894 & \cellcolor{purple!20}\textbf{5.913} & \cellcolor{yellow!20}\underline{64.981} & \cellcolor{yellow!20}\underline{9.932} & 15.843 & 8.757  & \underline{9.962} & \cellcolor{yellow!20}\textbf{9.753} & \underline{7.676} & \underline{14.639} \\
\textbf{AWQ}             & \textbf{multi10} & 7.688 & \underline{5.874} & \underline{5.926} & 65.773 & 9.994  & \underline{15.590} & \cellcolor{purple!20}\textbf{8.706} & 10.040 & \underline{9.788} & 7.695 & 14.707 \\
\hline
% ------- AWQ with code/math: only overall colors (no bold/underline) -------
\textbf{AWQ}             & \textbf{code-multi}      & \cellcolor{purple!20}7.663 & 5.901 & 5.918 & 65.792 & \cellcolor{yellow!20}9.932  & 15.675 & 8.770 & 9.982  & 9.762  & \cellcolor{yellow!20}7.648 & 14.704 \\
\textbf{AWQ}             & \textbf{math-multi}      & 7.674 & 5.889 & \cellcolor{yellow!20}5.915 & 65.154 & 9.934  & 15.574 & \cellcolor{yellow!20}8.718 & \cellcolor{yellow!20}9.951  & \cellcolor{purple!20}9.746  & 7.673 & \cellcolor{yellow!20}14.623 \\
\textbf{AWQ}             & \textbf{codemath-multi}  & \cellcolor{yellow!20}7.664 & 5.896 & 5.921 & 65.460 & 9.937  & \cellcolor{purple!20}15.499 & 8.730 & 10.001 & 9.811  & 7.683 & 14.660 \\
\textbf{AWQ}             & \textbf{codemath-multi10}& 7.675 & \cellcolor{yellow!20}5.873 & 5.918 & 65.944 & 9.975 & 15.679 & 8.745 & 10.048 & 9.790  & 7.711 & 14.736 \\ \hline \hline
\textbf{GPTQ}            & \textbf{en}      & 8.300 & 8.730 & 6.685 & 89.283 & 12.242 & 28.453 & 13.022 & 17.829 & 18.999 & 12.696 & 21.624 \\
\textbf{GPTQ}            & \textbf{sw}      & 9.286 & 9.633 & 7.908 & 95.637 & 15.054 & 35.093 & 15.035 & 19.274 & 23.086 & 18.786 & 24.879 \\
\textbf{GPTQ}            & \textbf{fr}      & 8.329 & 8.519 & 6.410 & 86.817 & 12.490 & 28.205 & 12.907 & 17.306 & 18.068 & 12.756 & 21.181 \\
\textbf{GPTQ}            & \textbf{xh}      & 8.612 & 7.844 & 6.693 & \underline{79.847} & 12.538 & 25.564 & 12.671 & \underline{15.714} & 17.028 & 11.990 & 19.850 \\
\textbf{GPTQ}            & \textbf{zh}      & 8.489 & 8.523 & 6.596 & 89.869 & 11.228 & 25.902 & 12.816 & 18.194 & 17.925 & 13.854 & 21.340 \\
\textbf{GPTQ}            & \textbf{multimix}& \underline{8.189} & 7.870 & 6.291 & \underline{81.481} & 11.220 & \underline{24.050} & 12.294 & 16.270 & 15.664 & 11.359 & 19.469 \\
\textbf{GPTQ}            & \textbf{multi}   & 8.211 & \underline{7.604} & \underline{6.260} & 82.564 & \underline{11.050} & 24.725 & \underline{11.890} & 16.304 & \underline{14.545} & \underline{10.060} & \underline{19.321} \\
\textbf{GPTQ}            & \textbf{multi10} & \textbf{8.184} & \cellcolor{yellow!20}\textbf{7.222} & \textbf{6.194} & \cellcolor{yellow!20}\textbf{77.444} & \textbf{10.787} & \cellcolor{yellow!20}\textbf{20.693} & \cellcolor{yellow!20}\textbf{11.329} & \cellcolor{yellow!20}\textbf{15.176} & \textbf{14.296} & \textbf{9.714} & \cellcolor{yellow!20}\textbf{18.104} \\
\hline
% ------- GPTQ with code/math: overall highlight only (no bold/underline) -------
\textbf{GPTQ}            & \textbf{code-multi}      & 8.191 & 7.496 & 6.214 & 79.968 & 10.950 & 22.969 & 11.547 & 15.836 & 13.895 & 9.699 & 18.676 \\
\textbf{GPTQ}            & \textbf{math-multi}      & 8.190 & 7.523 & \cellcolor{yellow!20}6.188 & 80.772 & 10.851 & 23.110 & 11.600 & 16.039 & 14.069 & 9.846 & 18.819 \\
\textbf{GPTQ}            & \textbf{codemath-multi}  & \cellcolor{yellow!20}8.174 & 7.495 & 6.193 & 80.949 & \cellcolor{yellow!20}10.744 & 22.026 & 11.610 & 16.203 & \cellcolor{yellow!20}13.838 & \cellcolor{yellow!20}9.534 & 18.677 \\
\textbf{GPTQ}            & \textbf{codemath-multi10}& \cellcolor{purple!20}8.159 & \cellcolor{purple!20}7.176 & \cellcolor{purple!20}6.183 & \cellcolor{purple!20}76.654 & \cellcolor{purple!20}10.695 & \cellcolor{purple!20}20.648 & \cellcolor{purple!20}11.237 & \cellcolor{purple!20}15.083 & \cellcolor{purple!20}13.803 & \cellcolor{purple!20}9.464 & \cellcolor{purple!20}17.910 \\

 \bottomrule
\end{tabular}
}
\caption{Llama3.1 8B perplexity on Wikipedia (lower is better). Natural-language calibration: \textbf{bold} = best, \underline{underline} = second-best. AWQ (incl.\ code/math): \textcolor{black}{\fcolorbox{white}{purple!20}{\strut}} best, \textcolor{black}{\fcolorbox{white}{yellow!20}{\strut}} second-best. AWQ benefits from language-matched calibration locally, whereas GPTQ prefers diverse multilingual calibration (\texttt{multi10}) for robust averages.}
\label{tab:Llama}
\end{table*}

\hl{\textbf{Argument 2}: Diverse Calibration Sets Account for Rare and Extreme Data.}

\begin{figure*}[htb!]
  \centering
  %--------------------------------------------------%
  \begin{subfigure}[b]{0.23\textwidth}
    \centering
    \includegraphics[width=\linewidth]{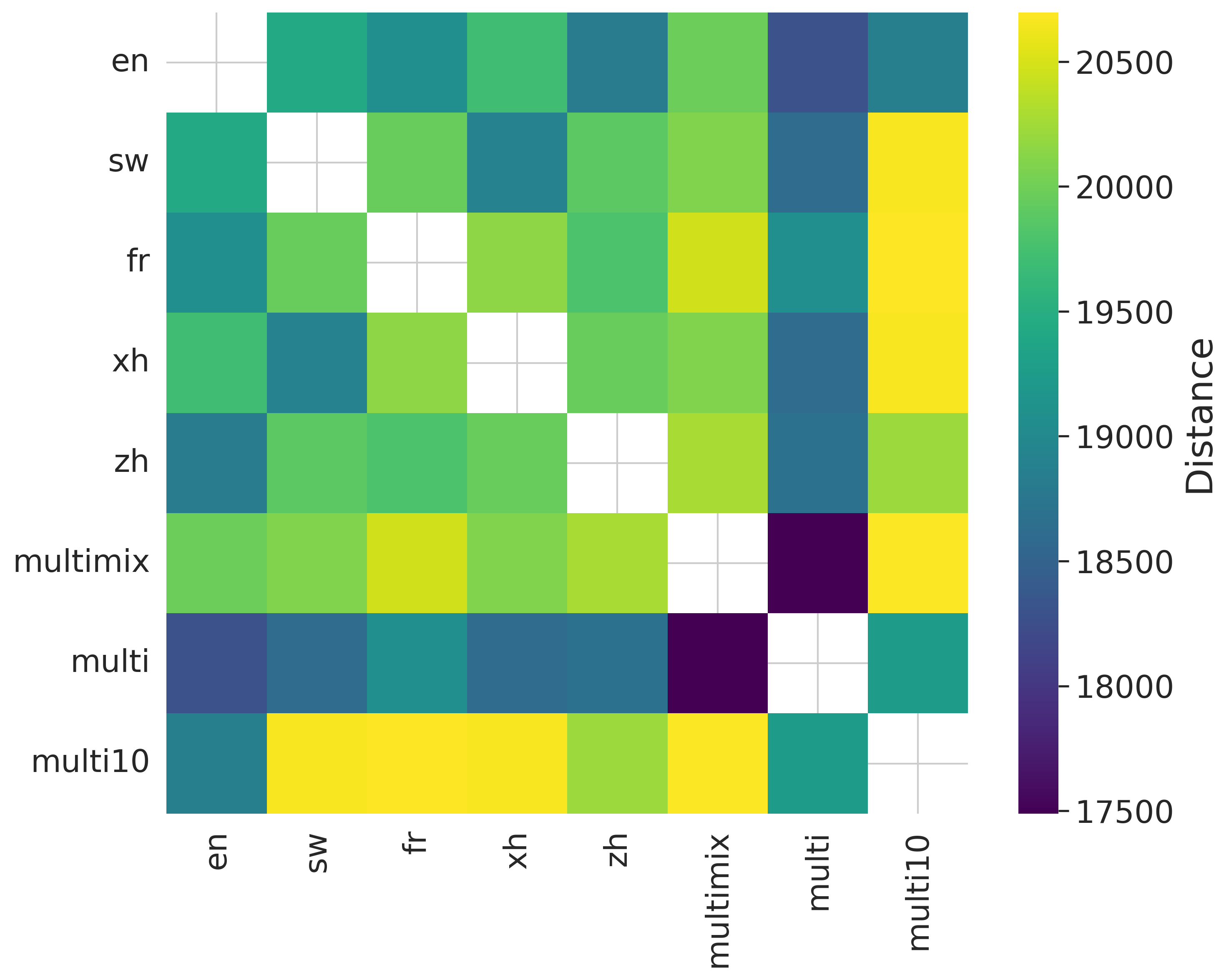}
    \caption{Layer 0 (Llama3.1 8B)}
    \label{fig:Llama_hess1}
  \end{subfigure}\hfill
  %--------------------------------------------------%
  \begin{subfigure}[b]{0.23\textwidth}
    \centering
    \includegraphics[width=\linewidth]{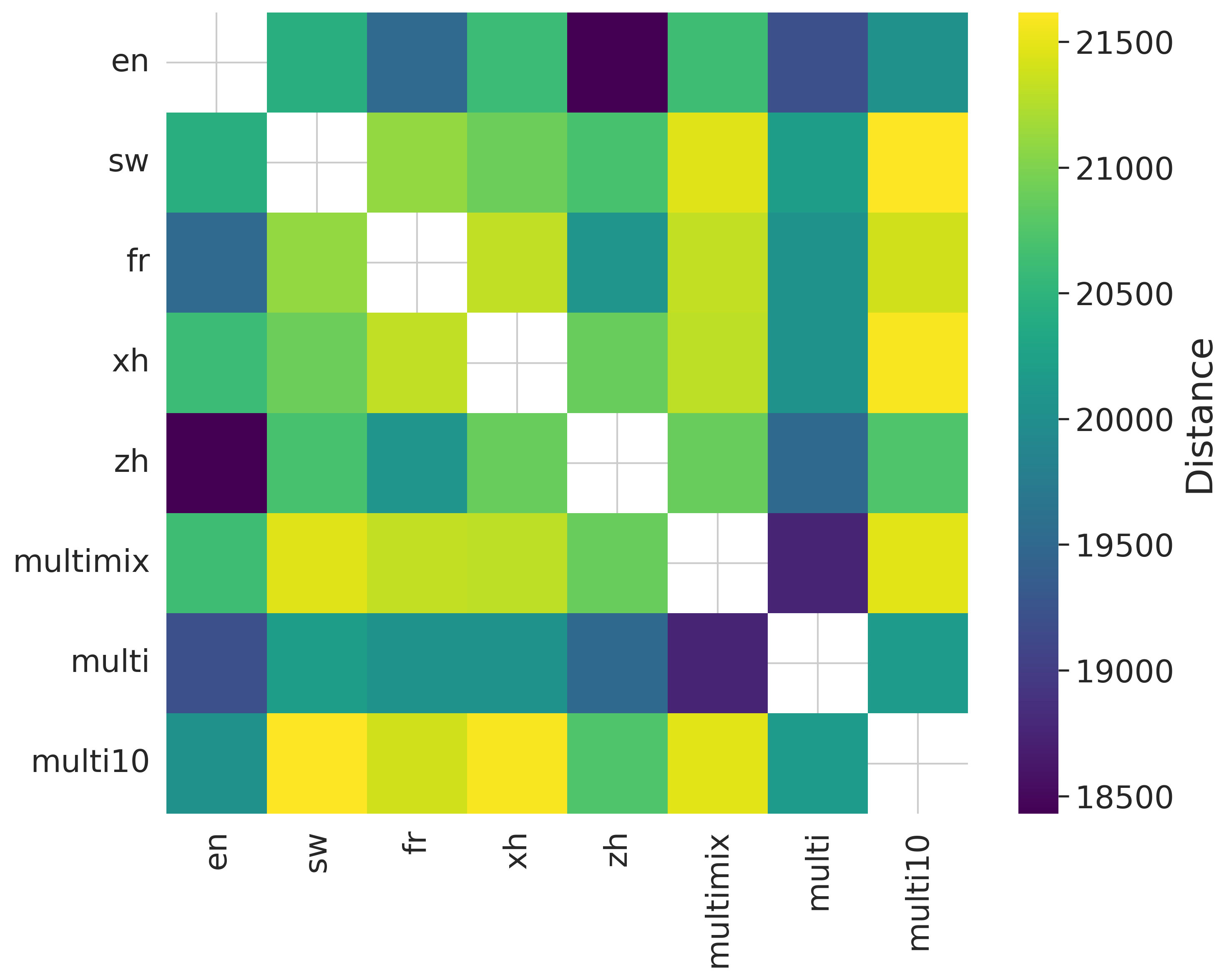}
    \caption{Layer 31 (Llama3.1 8B)}
    \label{fig:Llama_hess2}
  \end{subfigure}\hfill
  %--------------------------------------------------%
  \begin{subfigure}[b]{0.23\textwidth}
    \centering
    \includegraphics[width=\linewidth]{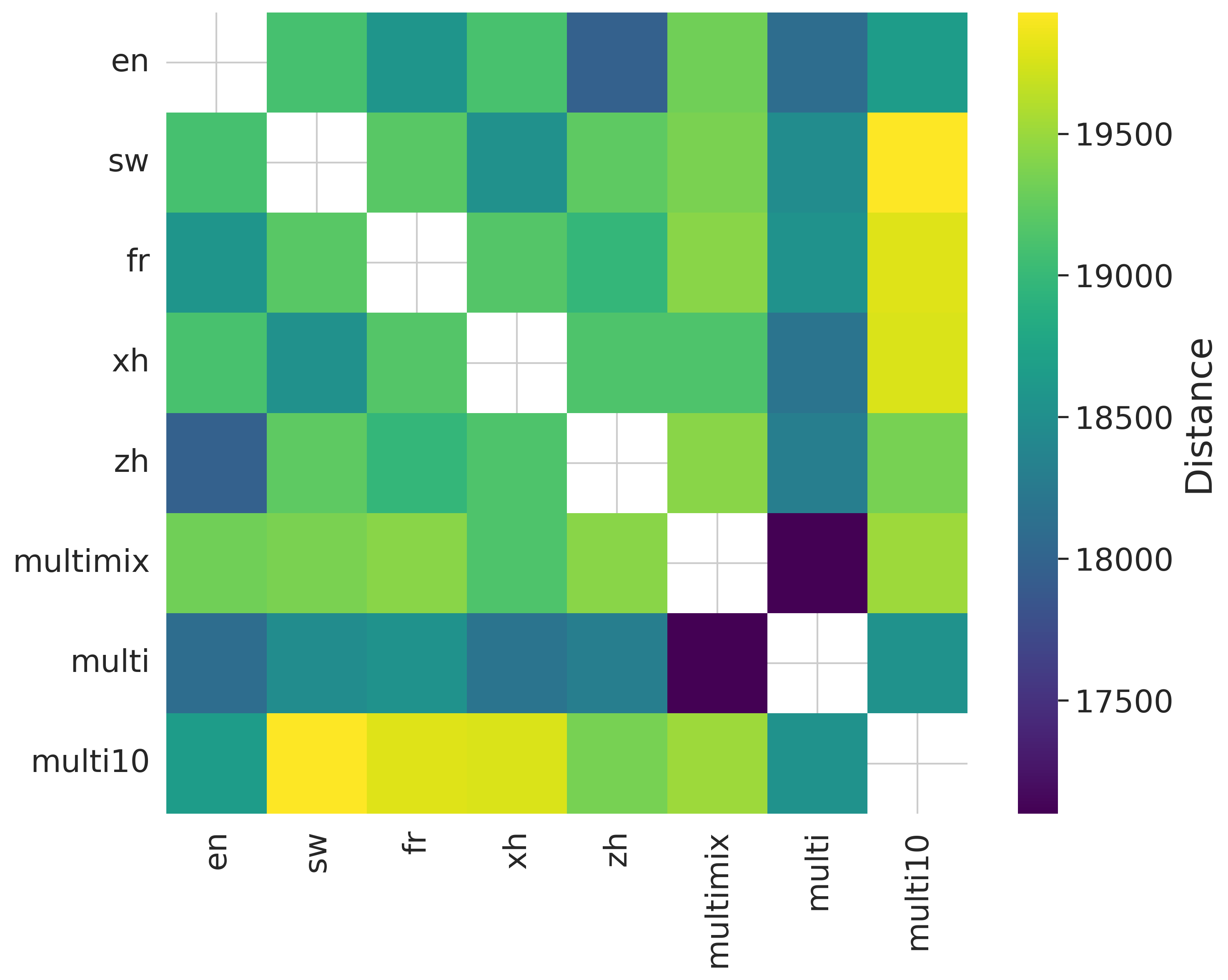}
    \caption{Layer 0 (Qwen2.5 7B)}
    \label{fig:qwen_hess1}
  \end{subfigure}\hfill
  %--------------------------------------------------%
  \begin{subfigure}[b]{0.23\textwidth}
    \centering
    \includegraphics[width=\linewidth]{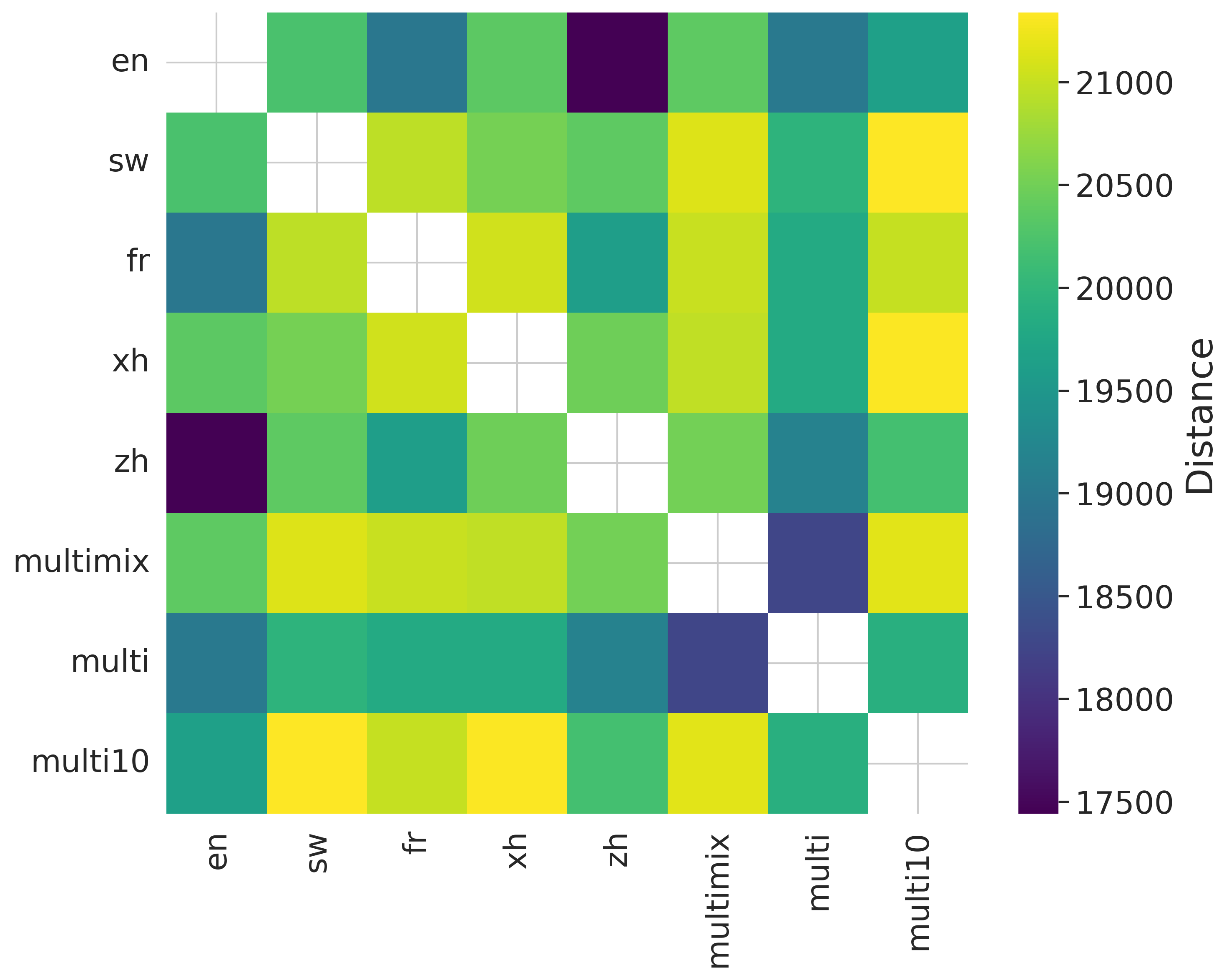}
    \caption{Layer 27 (Qwen2.5 7B)}
    \label{fig:qwen_hess2}
  \end{subfigure}
  %--------------------------------------------------%

  \caption{Heatmaps show inverse-Hessian distances (higher = larger change) between GPTQ calibrations for Llama3.1 8B (layers 0, 31) and Qwen2.5 7B (layers 0, 27). Among them, \texttt{multi10} shows the largest shifts, \texttt{multimix} moderate, and \texttt{multi} the smallest. Some monolingual pairs such as \texttt{en}–\texttt{zh} remain relatively close.}
  \label{fig:hessian_4panel}
\end{figure*}

\textit{Multilingual calibration sets offer the highest average performance among all settings.} On Llama3.1 8B (Wikipedia), the multilingual mixes (\texttt{multimix}, multi, and \texttt{multi10}) consistently sit at the top of the Avg column, with \texttt{math-multi} essentially tying or edging the best plain multilingual average; under GPTQ, \texttt{multi10} is the best plain mix and \texttt{codemath-multi10} is the best overall average, highlighting GPTQ’s sensitivity to broad coverage (see \Cref{tab:Llama}). All code/math variants are constructed by \emph{uniformly sampling} from the respective pools (code, math, or code+math) to fill the same calibration budget as other settings, with identical quantization hyperparameters—so the gain comes from \emph{content}, not data volume or tuning. We add code and math \emph{specifically} to test our rare/unique-token hypothesis: these domains contain high-rarity symbols, numerals, operators, and identifier-like subwords that broaden coverage beyond natural text. While AWQ occasionally rewards language-matched calibration with the very best per-language scores (e.g., for \texttt{xh}), the multilingual (+ code/math) variants remain the most reliable choice across languages. The same qualitative pattern holds for Qwen2.5 7B in the Appendix (\Cref{tab:QWEN})—multilingual \texttt{multi}/\texttt{multi10} rank at or near the top for both AWQ and GPTQ—reinforcing the general principle that broader, more diverse calibration improves average performance across models, datasets, and quantizers.

\begin{figure*}[htbp!]
  \centering
  %--------------------------------------------------%
  \begin{subfigure}[b]{0.23\textwidth}
    \centering
    \includegraphics[width=0.8\linewidth]{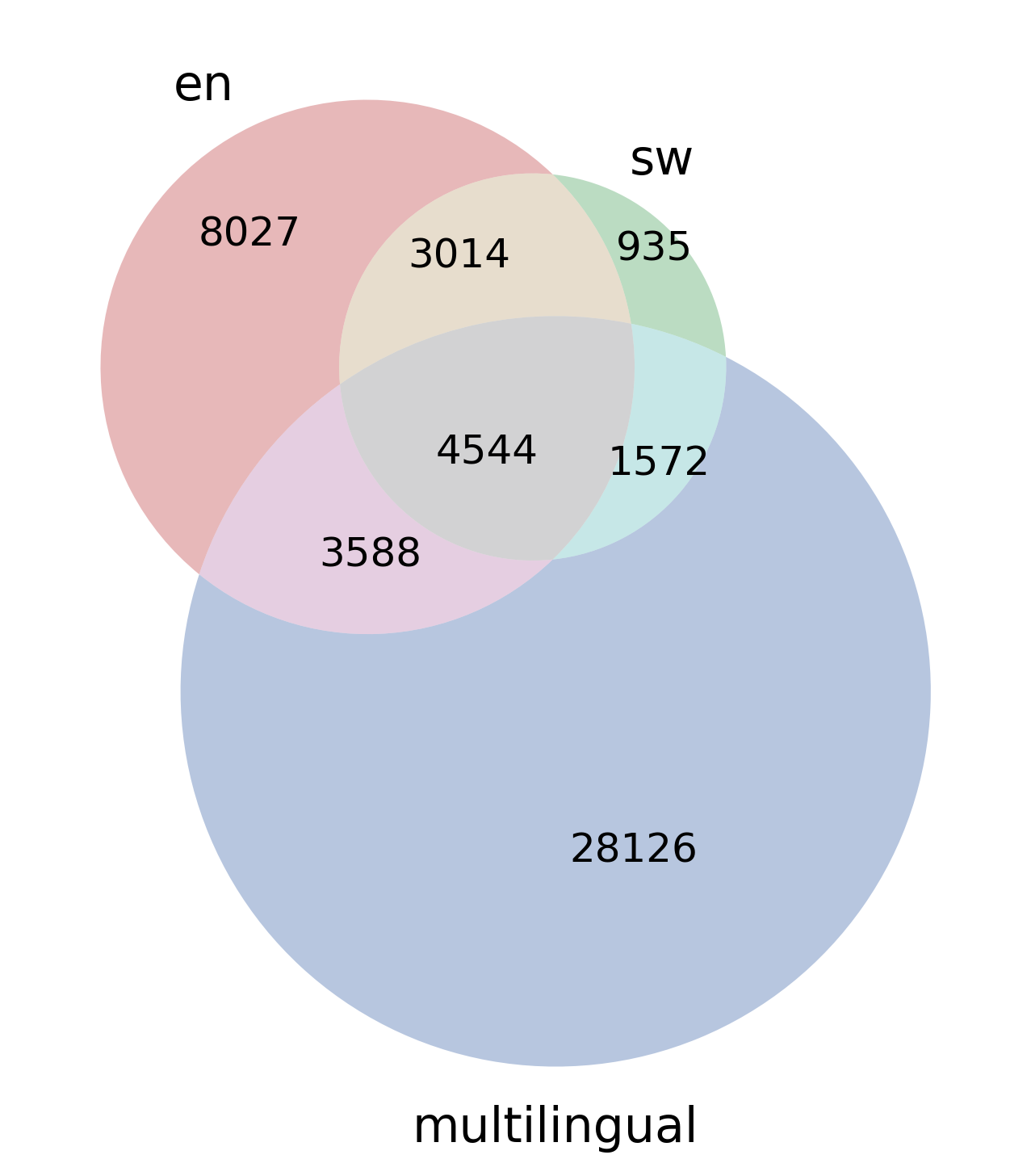}
    \caption{Llama3.1 8B}
    \label{fig:Llama_1}
  \end{subfigure}\hfill
  %--------------------------------------------------%
  \begin{subfigure}[b]{0.23\textwidth}
    \centering
    \includegraphics[width=0.8\linewidth]{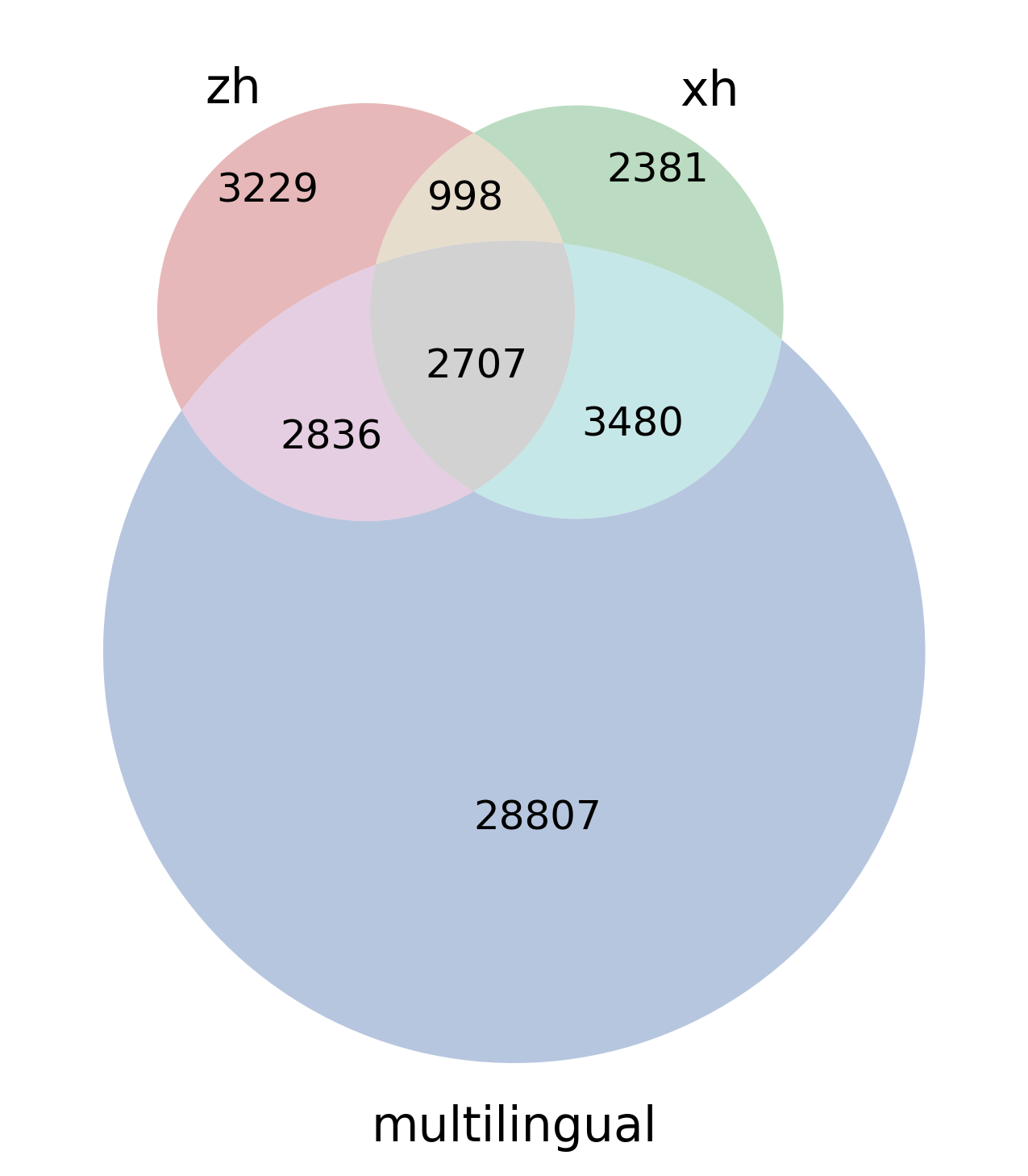}
    \caption{Llama3.1 8B}
    \label{fig:Llama_2}
  \end{subfigure}\hfill
  %--------------------------------------------------%
  \begin{subfigure}[b]{0.23\textwidth}
    \centering
    \includegraphics[width=0.8\linewidth]{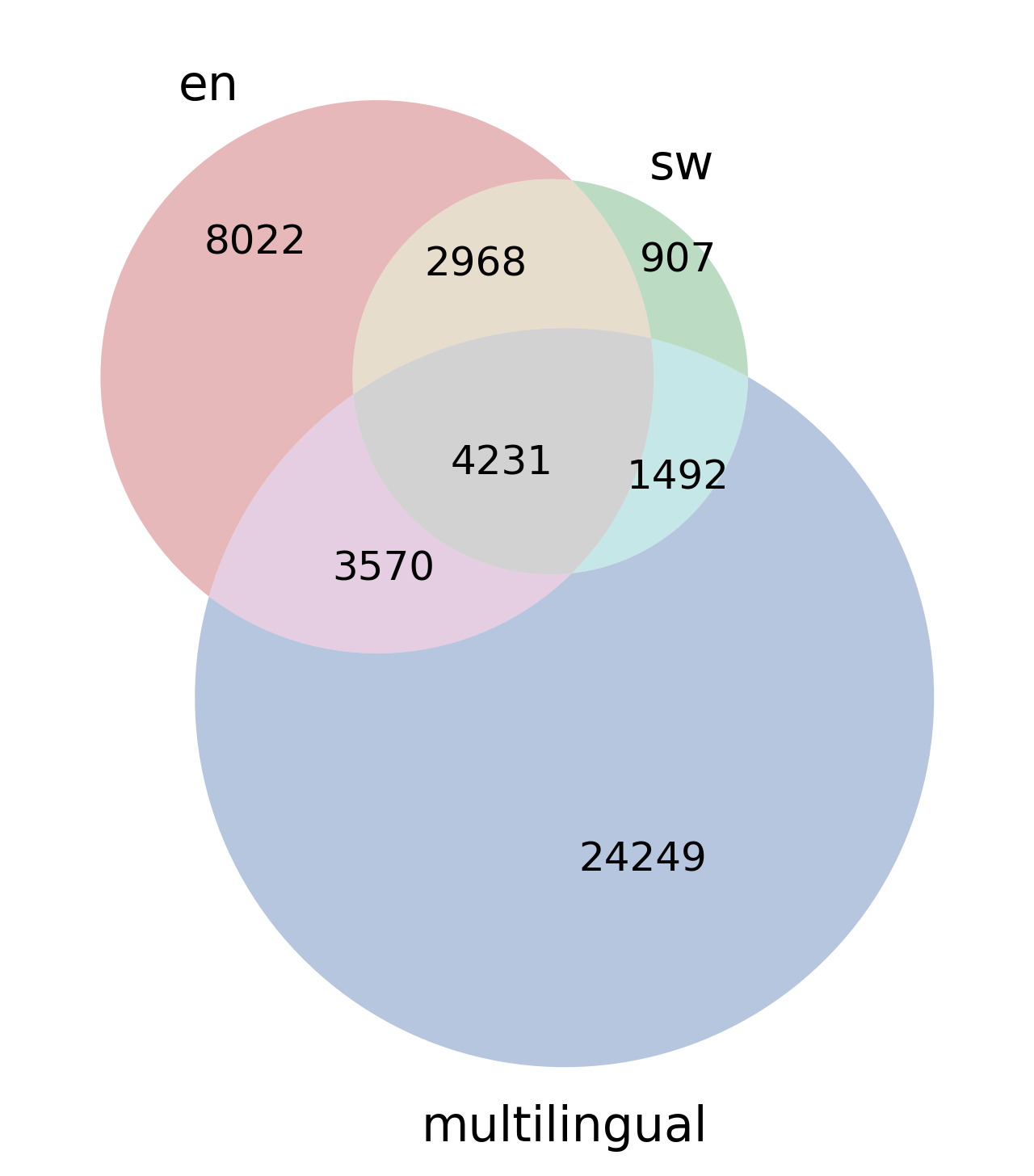}
    \caption{Qwen2.5 7B}
    \label{fig:qwen_1}
  \end{subfigure}\hfill
  %--------------------------------------------------%
  \begin{subfigure}[b]{0.23\textwidth}
    \centering
    \includegraphics[width=\linewidth]{img/vocab_overlap_llama2.png}
    \caption{Qwen2.5 7B}
    \label{fig:qwen_2}
  \end{subfigure}
  %--------------------------------------------------%

  \caption{Overlap analysis between the multilingual calibration set and single-language sets. Each language overlaps more with the multilingual set ($\approx$1.5k–3.6k activations) than with another single language ($\approx$1k–3k). The shared region across single languages and multilingual ($\approx$2.7k–4.5k) is also larger than direct language–language intersections. This shows that multilingual calibration covers a broader slice of the activation tail and reliably captures extremes that individual languages encounter at inference time.}
  \label{fig:overlap_4panel}
\end{figure*}

\textit{Broader token/structure coverage produces longer activation tails at calibration time, reducing clipping and quantization error.} Multilingual mixes exhibit the largest unique vocabularies and higher tokens per example (\Cref{fig:vocab}),\allowbreak{} exposing quantizers to more subwords, numerals, punctuation, and script-specific patterns. Their vocabularies also overlap more with each single language---$\approx$1.5k--3.6k shared types vs.\ $\approx$1k--3k for direct language--to--language pairs---\allowbreak{} and three-way overlaps are larger when a multilingual set is included (\Cref{fig:overlap_4panel}).\allowbreak{} This richer lexical/structural diversity yields heavier upper activation tails (\Cref{fig:multiact}),\allowbreak{} the very region where under-calibrated scales would otherwise clip unseen extremes. Adding code/math amplifies this effect by injecting rare symbols and bracketed/numeric patterns,\allowbreak{} further extending the tails and lowering post-quantization error---which is exactly what we observe in \Cref{tab:Llama}.

\hl{\textbf{Argument 3}: Calibration effectiveness is quantizer-dependent: the same calibration set can help or hurt differently depending on the algorithm’s mechanics.}

In \Cref{fig:Delta}, GPTQ shows large swings across calibration sets (up to \(3.52\) ppl), while AWQ varies only slightly (max \(\approx0.35\) ppl), echoing our claim that calibration is not language-agnostic and must be chosen with intent. Concretely, multilingual mixes (especially \texttt{multi10}) deliver the most robust averages for GPTQ, whereas AWQ’s gains are steadier and often maximized by language-matched calibration for a given target language.

To understand this difference, we examine how each quantization algorithm interacts with the calibration data at specific layers. In particular, we focus on the extremity layers — those most affected by quantization noise. For Llama3.1 8B, these are layers 0 and 31; for Qwen2.5 7B, layers 0 and 27. These layers consistently exhibit the highest quantization error magnitudes (see \Cref{sec:appendix}), and thus provide a focused lens through which to study how calibration data influences quantization behavior.

\emph{AWQ’s activation scaling keeps channel identity stable, so calibration language mainly adjusts magnitudes, not which channels matter.} In \Cref{fig:channel}, we plot the maximum per-channel activations at layer 31 (gate and query projections). Across all calibration languages, AWQ consistently selects the same “salient” channel, but the absolute activation peaks differ. This explains AWQ’s modest perplexity deltas: when channel identity is stable, language choice fine-tunes the rescaling but rarely causes large distributional mismatches. Practically, if you must optimize a single language with AWQ, match the calibration language; if you need cross-language robustness, multilingual remains near-optimal.

\emph{GPTQ’s Hessian-based error compensation is acutely sensitive to the calibration distribution, but diverse calibration stabilizes it.} Because GPTQ propagates quantization residuals using an inverse-Hessian estimated from calibration activations, changing the calibration language changes the second-order statistics the algorithm relies on. \Cref{fig:hessian_4panel} visualizes large inverse-Hessian differences across single-language sets, aligning with GPTQ’s larger $\Delta$ ppl swings; by contrast, \texttt{multi} calibrations cluster tightly in this space, indicating that linguistically diverse calibration dampens GPTQ’s sensitivity. These results support our core claim: robust quantization needs linguistically diverse calibration. 

To further test quantizer generality, we additionally evaluate Any4, a recent activation-aware post-training quantizer, on Qwen~2.5 (see Appendix~\ref{sec:any4}). Despite its different mechanics, Any4 exhibits the same qualitative behavior: multilingual and multilingual-mixture calibration consistently yield lower average perplexity across languages, while monolingual calibration provides only localized improvements.
This confirms that the benefits of language-diverse calibration are not specific to GPTQ or AWQ, but reflect a general interaction between calibration data and activation distributions across quantization families.

Practically, we recommend balanced multilingual calibration for GPTQ and language-matched calibration (with a multilingual fallback) for AWQ. Understanding these algorithm-specific interactions is essential for selecting calibration data that best synergizes with each quantizer’s mechanics.

\section{Conclusion and Future Work}
Effective quantization is essential for deploying multilingual LLMs in GPU-poor environments. We show that appropriate calibration is a valuable tool to minimise loss of performance. In this paper, we presented the first systematic study of multilingual calibration sets for LLM quantization, challenging the convention of using English-only calibration sets. 
We observed that non‐English and mixed‐language calibration sets consistently outperform the English‐only baseline, achieving up to a \(3.52\) perplexity reduction on Llama3.1 using GPTQ. English‐only calibration leaves substantial accuracy gaps for many target languages. We also demonstrate that multilingual calibration sets exhibit longer activation tails (\Cref{fig:multiact}), indicating better coverage of outlier tokens and activation values. This broader coverage directly translates into lower quantization error and more robust performance across languages. Lastly, we show that algorithm mechanics matter with GPTQ’s Hessian‐based error compensation being more sensitive to the calibration language choice than AWQ’s activation‐aware scaling (\Cref{fig:Delta}). This interaction implies that calibration data must be chosen in harmony with the quantizer’s algorithmic design.

These findings demonstrate that \emph{calibration set design} is a pivotal factor in preserving multilingual LLM accuracy under low‐precision constraints. From our experiments, we distill three practical guidelines:
\begin{itemize}
  \item Use a single multilingual calibration set (e.g., multi10) for general multilingual deployment \allowbreak{} and reserve language-matched calibration only for single-language–dominated systems,\allowbreak{} as an optional optimization rather than a requirement.
  \item Targeted AWQ: If you are optimizing for one language, choose a matching calibration language dataset, with a multilingual fallback for broader robustness.  
  \item Targeted GPTQ: Prefers balanced multilingual calibration; adding special tokens (code/math) can further stabilize second-order statistics. 
\end{itemize}

\section*{Limitations}

In this work, we focus on varying the language aspect of a calibration set, evaluating on perplexity. However, evaluating only perplexity provides a limited view of real-world utility; extending our framework to downstream tasks e.g., translation would more directly quantify the practical benefits of non-English and multilingual calibration. We also limit our analysis to 10 languages, two LLMs (Llama3.1 8B, Qwen2.5 7B), and two post-training quantization methods (GPTQ, AWQ) due to computational resource limits; broadening this scope to include additional languages, model families, model sizes, and quantizers is warranted to strengthen our argument. Lastly, we did not explore dynamic calibration sample selection based on quantization error signals; future work should explore this area which may lead to further reduction in quantization loss.

\section*{Acknowledgments}

We thank Benjamin Rosman, Jade Abbott and the anonymous reviewers for the comments and contributions to the draft.

\bibliography{anthology_1,anthology_2,custom}

\newpage
\appendix
\section{Appendix}
\label{sec:appendix}

\subsection{Cross-Model Check on Wikipedia (Qwen2.5 7B)}

\begin{table*}[htbp!]
\centering
\resizebox{\textwidth}{!}{
\begin{tabular}{l|l|rrrrrrrrrrr}
\textbf{Quantization} & \textbf{Calibration} &
  \multicolumn{1}{l}{\textbf{en}} &
  \multicolumn{1}{l}{\textbf{sw}} &
  \multicolumn{1}{l}{\textbf{fr}} &
  \multicolumn{1}{l}{\textbf{xh}} &
  \multicolumn{1}{l}{\textbf{zh}} &
  \multicolumn{1}{l}{\textbf{st}} &
  \multicolumn{1}{l}{\textbf{yo}} &
  \multicolumn{1}{l}{\textbf{zu}} &
  \multicolumn{1}{l}{\textbf{ha}} &
  \multicolumn{1}{l}{\textbf{ig}} &
  \multicolumn{1}{l}{\textbf{Avg}} \\ \toprule
\textbf{FP16}          & --               & 6.835 & 13.163 & 6.227 & 20.296 & 9.948 & 33.537 & 12.608 & 16.983 & 24.070 & 19.754 & 16.642 \\
\textbf{4-bit}         & --               & 7.562 & 14.766 & 6.639 & 21.599 & 10.769 & 35.509 & 13.723 & 18.828 & 26.601 & 21.456 & 17.845 \\ \hline
\textbf{AWQ}           & \textbf{en}       & \cellcolor{purple!20}\textbf{7.212} & 14.097 & 6.436 & 20.928 & 10.594 & 33.497 & 12.828 & 17.398 & 25.240 & 20.663 & 16.689 \\
\textbf{AWQ}           & \textbf{sw}       & 7.267 & \cellcolor{yellow!20}\underline{13.793} & 6.462 & 20.737 & 10.592 & 33.569 & \cellcolor{yellow!20}\underline{12.737} & 17.194 & 25.275 & 20.754 & 16.638 \\
\textbf{AWQ}           & \textbf{fr}       & \cellcolor{yellow!20}\underline{7.231} & 14.132 & \cellcolor{yellow!20}\underline{6.378} & 20.946 & 10.579 & 34.085 & 12.797 & 17.432 & 25.554 & 21.137 & 16.727 \\
\textbf{AWQ}           & \textbf{xh}       & 7.235 & \cellcolor{purple!20}\textbf{13.752} & 6.427 & \cellcolor{yellow!20}\underline{20.701} & 10.635 & 33.428 & 12.771 & \cellcolor{purple!20}\textbf{17.055} & \cellcolor{yellow!20}\underline{25.183} & 20.695 & 16.598 \\
\textbf{AWQ}           & \textbf{zh}       & 7.235 & 13.902 & 6.456 & 20.781 & \cellcolor{purple!20}\textbf{10.467} & \cellcolor{purple!20}\textbf{32.996} & \cellcolor{purple!20}\textbf{12.637} & 17.267 & 25.289 & 20.716 & \cellcolor{yellow!20}\underline{16.475} \\
\textbf{AWQ}           & \textbf{multimix} & 7.243 & 13.900 & 6.415 & 20.743 & \cellcolor{yellow!20}\underline{10.563} & 33.173 & 12.738 & 17.279 & \cellcolor{purple!20}\textbf{25.123} & \cellcolor{yellow!20}\underline{20.611} & 16.479 \\
\textbf{AWQ}           & \textbf{multi}    & \cellcolor{purple!20}\textbf{7.212} & 13.842 & \cellcolor{purple!20}\textbf{6.350} & \cellcolor{purple!20}\textbf{20.599} & 10.568 & \cellcolor{yellow!20}\underline{32.997} & \cellcolor{purple!20}\textbf{12.637} & \cellcolor{yellow!20}\underline{17.088} & 25.213 & \cellcolor{purple!20}\textbf{20.448} & \cellcolor{purple!20}\textbf{16.395} \\ \hline\hline
\textbf{GPTQ}          & \textbf{en}       & \cellcolor{purple!20}\textbf{7.324} & 14.958 & 6.617 & 21.665 & 11.661 & 37.648 & 13.779 & 18.789 & 26.789 & 21.717 & 17.995 \\
\textbf{GPTQ}          & \textbf{sw}       & 7.570 & \cellcolor{purple!20}\textbf{13.703} & 6.677 & \cellcolor{yellow!20}\underline{20.996} & 11.805 & 35.089 & 13.286 & 17.898 & \cellcolor{yellow!20}\underline{25.020} & 20.816 & 17.186 \\
\textbf{GPTQ}          & \textbf{fr}       & 7.489 & 14.715 & \cellcolor{yellow!20}\underline{6.453} & 21.516 & 11.772 & 36.549 & 13.606 & 18.705 & 26.928 & 22.139 & 17.707 \\
\textbf{GPTQ}          & \textbf{xh}       & 7.648 & 13.974 & 6.770 & \cellcolor{purple!20}\textbf{20.627} & 11.650 & \cellcolor{purple!20}\textbf{34.379} & 13.439 & \cellcolor{purple!20}\textbf{17.343} & 25.395 & 21.158 & 17.178 \\
\textbf{GPTQ}          & \textbf{zh}       & 7.473 & 14.973 & 6.683 & 21.543 & 11.689 & 36.945 & 13.767 & 18.705 & 27.305 & 21.935 & 17.802 \\
\textbf{GPTQ}          & \textbf{multimix} & 7.451 & 14.385 & 6.562 & 21.047 & 13.147 & 35.451 & 13.335 & 18.007 & 25.605 & 21.935 & 17.494 \\
\textbf{GPTQ}          & \textbf{multi}    & 7.482 & 14.121 & 6.465 & 21.126 & \cellcolor{yellow!20}\underline{10.712} & 34.550 & \cellcolor{yellow!20}\underline{13.134} & 17.994 & \cellcolor{purple!20}\textbf{24.841} & \cellcolor{purple!20}\textbf{20.523} & \cellcolor{yellow!20}\underline{16.895} \\
\textbf{GPTQ}          & \textbf{multi10}  & \cellcolor{yellow!20}\underline{7.424} & \cellcolor{yellow!20}\underline{13.732} & \cellcolor{purple!20}\textbf{6.451} & \cellcolor{purple!20}\textbf{20.627} & \cellcolor{purple!20}\textbf{10.520} & \cellcolor{yellow!20}\underline{34.390} & \cellcolor{purple!20}\textbf{12.976} & \cellcolor{yellow!20}\underline{17.424} & 25.378 & \cellcolor{yellow!20}\underline{20.616} & \cellcolor{purple!20}\textbf{16.563} \\ \bottomrule
\end{tabular}
}
\caption{Per-language perplexity for Qwen2.5 7B (lower is better) on the Wikipedia evaluation set, with per-language scores and the overall \textbf{Avg}. The first two rows show the FP16 baseline and an uncalibrated 4-bit quantization result. The remaining rows report results using AWQ (top block) and GPTQ (bottom block) with calibration data drawn from a single language (\texttt{en}, \texttt{sw}, \texttt{fr}, \texttt{xh}, \texttt{zh}) or from one of three multilingual calibration sets (\texttt{multimix}, \texttt{multi}, \texttt{multi10}). Bold values\textcolor{black}{\fcolorbox{white}{purple!20}{\strut}} indicate the best perplexity per language, while underlined\textcolor{black}{\fcolorbox{white}{yellow!20}{\strut}} values indicate the second-best.  For AWQ, the multilingual \texttt{multi} configuration yields the best or second-best performance in 7 out of 10 languages, outperforming most language-specific variants and suggesting a strong generalization benefit from multilingual calibration. For GPTQ, \texttt{multi10} delivers the best results in 5 out of 10 languages and is second-best in 4 others, demonstrating that diverse calibration data offers consistent improvements across languages, but not as uniformly as in AWQ.}
\label{tab:QWEN}
\end{table*}

On a different model (Qwen2.5 7B on C4), multilingual calibration again delivers the strongest averages. \Cref{tab:QWEN} mirrors the Llama–Wikipedia trend: multilingual \texttt{multi}/\texttt{multi10} rank at or near the top for both AWQ and GPTQ, while some language-matched settings win per-language minima. This cross-model replication strengthens our main claim that broad linguistic coverage in calibration improves average robustness.

\subsection{Additional Model Family: BLOOM}
\label{sec:appendix_bloom}

To further test whether our findings depend on English-dominant pretraining,
we also evaluate BLOOMZ-7B1-MT, a multilingual model trained with a more globally
balanced pretraining mixture and weaker English dominance than Llama~3.1 or Qwen~2.5.
Table~\ref{tab:bloom_results} reports AWQ perplexity results for BLOOMZ-7B1-MT across
calibration strategies. We observe the same qualitative pattern as in earlier experiments:
language-matched calibration yields localized improvements for the corresponding
or closely related languages, while multilingual calibration remains competitive
on average. These results indicate that the benefits of multilingual and language-aware
calibration extend beyond English-centric pretraining regimes and are not an
artifact of a single model family.

\begin{table}[htbp]
\centering
\resizebox{\columnwidth}{!}{
\begin{tabular}{lccccc}
\toprule
\textbf{Calibration} & \textbf{en} & \textbf{sw} & \textbf{zh} & \textbf{xh} & \textbf{st} \\
\midrule
\textbf{en} & \cellcolor{purple!20}\textbf{14.3781} & 18.2803 & 22.9508 & \cellcolor{yellow!20}\underline{120.1768} & \cellcolor{purple!20}\textbf{37.4696} \\
\textbf{sw} & 14.4567 & \cellcolor{purple!20}\textbf{18.1266} & 22.9426 & \cellcolor{purple!20}\textbf{119.5649} & 37.7456 \\
\textbf{xh} & 14.4829 & 18.2722 & 22.9723 & 120.3146 & 38.2460 \\
\textbf{zh} & 14.4376 & 18.3402 & \cellcolor{purple!20}\textbf{22.8780} & 120.7325 & \cellcolor{yellow!20}\underline{37.5867} \\
\textbf{multi10} & 14.4096 & \cellcolor{yellow!20}\underline{18.1300} & 22.9440 & 120.4858 & 37.8344 \\
\textbf{multimix} & \cellcolor{yellow!20}\underline{14.4057} & 18.2377 & \cellcolor{yellow!20}\underline{22.9303} & 120.5694 & 37.6794 \\
\textbf{multi} & 14.5457 & 18.3571 & 22.9825 & 120.9074 & 37.9532 \\
\bottomrule
\end{tabular}}
\caption{AWQ perplexity results for BLOOMZ-7B1-MT under different calibration strategies.
Language-matched calibration provides localized gains, while multilingual calibration
remains competitive across languages.}
\label{tab:bloom_results}
\end{table}

\subsection{Additional Downstream Results}
\label{sec:appendix_downstream}

Table~\ref{tab:appendix_full_downstream} reports full language-specific downstream results
for Llama3.1 8B using GPTQ quantization across XNLI,
XStoryCloze, and Global MMLU.

\begin{table*}[htbp]
\centering
\small
\resizebox{\textwidth}{!}{%
\begin{tabular}{l|ccccc|cccc|ccccc}
\toprule
& \multicolumn{5}{c}{\textbf{XNLI}} 
& \multicolumn{4}{c}{\textbf{XStoryCloze}} 
& \multicolumn{5}{c}{\textbf{Global MMLU}} \\
\cmidrule(lr){2-6} \cmidrule(lr){7-10} \cmidrule(lr){11-15}
\textbf{Calibration}
& \textbf{en} & \textbf{fr} & \textbf{sw} & \textbf{zh} & \textbf{hi}
& \textbf{en} & \textbf{sw} & \textbf{zh} & \textbf{hi}
& \textbf{en} & \textbf{sw} & \textbf{fr} & \textbf{xh} & \textbf{yo} \\
\midrule
\textbf{en}
& \textbf{53.94 $\pm$ 1.00} & 51.81 $\pm$ 1.00 & 38.11 $\pm$ 0.97 & \underline{39.20 $\pm$ 0.98} & 44.50 $\pm$ 1.00
& 78.62 $\pm$ 1.06 & 53.74 $\pm$ 1.28 & \underline{65.19 $\pm$ 1.23} & 62.34 $\pm$ 1.25
& 64.00 $\pm$ 2.39 & 37.75 $\pm$ 2.42 & 57.50 $\pm$ 2.47 & 53.75 $\pm$ 2.48 & 31.50 $\pm$ 2.31 \\

\textbf{fr}
& \underline{53.29 $\pm$ 1.00} & 52.45 $\pm$ 1.00 & 37.55 $\pm$ 0.97 & 37.07 $\pm$ 0.97 & \underline{46.79 $\pm$ 1.00}
& 78.69 $\pm$ 1.05 & \underline{54.93 $\pm$ 1.28} & 65.06 $\pm$ 1.23 & 62.74 $\pm$ 1.24
& 63.00 $\pm$ 2.39 & \underline{39.00 $\pm$ 2.44} & 55.50 $\pm$ 2.49 & 51.00 $\pm$ 2.49 & 32.75 $\pm$ 2.32 \\

\textbf{sw}
& 52.45 $\pm$ 1.00 & 52.05 $\pm$ 1.00 & \textbf{40.20 $\pm$ 0.98} & \textbf{39.56 $\pm$ 0.98} & 46.67 $\pm$ 1.00
& 77.70 $\pm$ 1.07 & 53.01 $\pm$ 1.28 & 62.74 $\pm$ 1.24 & 58.50 $\pm$ 1.27
& 61.00 $\pm$ 2.38 & 37.75 $\pm$ 2.43 & 44.00 $\pm$ 2.48 & 50.00 $\pm$ 2.49 & 28.75 $\pm$ 2.26 \\

\textbf{zh}
& 53.01 $\pm$ 1.00 & \textbf{53.01 $\pm$ 1.00} & 38.76 $\pm$ 0.98 & 35.30 $\pm$ 0.96 & 45.66 $\pm$ 1.00
& \textbf{79.81 $\pm$ 1.03} & 54.00 $\pm$ 1.28 & 64.59 $\pm$ 1.23 & 63.53 $\pm$ 1.24
& \textbf{66.50 $\pm$ 2.35} & 37.75 $\pm$ 2.44 & 56.50 $\pm$ 2.46 & 53.25 $\pm$ 2.49 & 32.25 $\pm$ 2.32 \\

\textbf{xh}
& 51.93 $\pm$ 1.00 & 52.41 $\pm$ 1.00 & 39.48 $\pm$ 0.98 & 37.91 $\pm$ 0.97 & 44.18 $\pm$ 1.00
& 78.49 $\pm$ 1.06 & 54.60 $\pm$ 1.28 & 64.26 $\pm$ 1.23 & 62.28 $\pm$ 1.25
& 62.50 $\pm$ 2.41 & \underline{39.00 $\pm$ 2.45} & 49.75 $\pm$ 2.48 & 51.50 $\pm$ 2.50 & 33.00 $\pm$ 2.35 \\

\midrule
\textbf{multi}
& 52.97 $\pm$ 1.00 & 51.77 $\pm$ 1.00 & \underline{39.92 $\pm$ 0.98} & 37.43 $\pm$ 0.97 & 45.62 $\pm$ 1.00
& 78.76 $\pm$ 1.05 & 54.73 $\pm$ 1.28 & 64.73 $\pm$ 1.23 & \textbf{64.00 $\pm$ 1.24}
& \underline{66.25 $\pm$ 2.34} & \underline{39.00 $\pm$ 2.43} & 56.75 $\pm$ 2.46 & \textbf{56.00 $\pm$ 2.49} & \textbf{34.25 $\pm$ 2.35} \\

\textbf{multi10}
& 52.89 $\pm$ 1.00 & 50.64 $\pm$ 1.00 & 39.16 $\pm$ 0.98 & 36.99 $\pm$ 0.97 & \textbf{47.75 $\pm$ 1.00}
& \underline{78.89 $\pm$ 1.05} & \textbf{55.72 $\pm$ 1.28} & \textbf{65.32 $\pm$ 1.22} & 63.34 $\pm$ 1.24
& 65.00 $\pm$ 2.37 & \textbf{42.25 $\pm$ 2.46} & \underline{58.25 $\pm$ 2.46} & 53.75 $\pm$ 2.50 & 31.75 $\pm$ 2.31 \\

\textbf{multimix}
& 53.17 $\pm$ 1.00 & \underline{52.85 $\pm$ 1.00} & 38.92 $\pm$ 0.98 & 34.38 $\pm$ 0.95 & 45.50 $\pm$ 1.00
& 78.16 $\pm$ 1.06 & 53.47 $\pm$ 1.28 & 64.86 $\pm$ 1.23 & \underline{63.73 $\pm$ 1.24}
& 65.50 $\pm$ 2.37 & 36.50 $\pm$ 2.40 & \textbf{60.00 $\pm$ 2.44} & \underline{54.50 $\pm$ 2.49} & \underline{33.25 $\pm$ 2.31} \\
\bottomrule
\end{tabular}
}
\caption{Full language-specific downstream accuracy (\%) for Llama3.1 8B.
The table reports per-language results for XNLI (en, fr, sw, zh, hi), XStoryCloze (en, sw, zh, hi),
and Global MMLU (en, sw, fr, xh, yo).}
\label{tab:appendix_full_downstream}
\end{table*}

\subsection{Additional Quantizer: Any4}
\label{sec:any4}

To assess whether our findings extend beyond GPTQ and AWQ, we additionally evaluate Any4 \cite{pmlr-v267-elhoushi25a}, a recent activation-aware post-training quantizer, on Qwen~2.5 under 4-bit quantization. Table~\ref{tab:any4_results} reports language-specific perplexity across calibration strategies. We observe the same pattern seen for GPTQ and AWQ: multilingual and multilingual-mixture calibration sets achieve the lowest average perplexity across languages, while single-language calibration primarily yields localized gains. The relative ordering of calibration strategies remains stable, confirming that language-diverse calibration improves global robustness even under a newer, activation-aware quantization algorithm.

\begin{table*}[htbp]
\centering
\small
\resizebox{\textwidth}{!}{%
\begin{tabular}{l|l|cccccccccc}
\toprule
\textbf{Quant.} & \textbf{Calibration} & \textbf{en} & \textbf{sw} & \textbf{fr} & \textbf{xh} & \textbf{st} & \textbf{yo} & \textbf{zu} & \textbf{ha} & \textbf{ig} & \textbf{Avg} \\
\midrule
\textbf{FP16} & \textbf{None} & 9.29 & 28.69 & 8.10 & 53.91 & 38.93 & 20.88 & 40.20 & 36.64 & 22.89 & 28.84 \\
\textbf{4-bit} & \textbf{None} & 10.06 & 32.68 & 8.85 & 59.99 & 41.96 & 22.78 & 44.36 & 39.54 & 24.59 & 31.64 \\
\hline 
\textbf{Any4} & \textbf{en} 
& \cellcolor{purple!20}\textbf{9.82} 
& 31.77 
& \cellcolor{yellow!20}\underline{8.65} 
& 58.27 
& 41.14 
& 22.12 
& 43.25 
& 38.71 
& \cellcolor{yellow!20}\underline{24.15} 
& 30.88 \\

\textbf{Any4} & \textbf{sw} 
& 9.89 
& \cellcolor{purple!20}\textbf{31.57} 
& 8.66 
& 59.16 
& 41.37 
& 22.19 
& 43.75 
& 38.97 
& 24.55 
& 31.12 \\

\textbf{Any4} & \textbf{fr} 
& \cellcolor{yellow!20}\underline{9.84} 
& 31.80 
& \cellcolor{yellow!20}\underline{8.65} 
& \cellcolor{purple!20}\textbf{58.03} 
& 41.44 
& 22.19 
& 43.35 
& 38.86 
& 24.36 
& 30.95 \\

\textbf{Any4} & \textbf{xh} 
& 9.91 
& \cellcolor{yellow!20}\underline{31.59} 
& 8.68 
& \cellcolor{yellow!20}\underline{58.12} 
& 41.29 
& \cellcolor{yellow!20}\underline{21.96} 
& \cellcolor{yellow!20}\underline{43.19} 
& \cellcolor{purple!20}\textbf{38.53} 
& 24.29 
& \cellcolor{purple!20}\textbf{30.84} \\

\textbf{Any4} & \textbf{zh} 
& 9.88 
& 32.04 
& 8.68 
& 58.47 
& 41.82 
& 22.22 
& 43.51 
& 38.99 
& 24.39 
& 31.11 \\

\textbf{Any4} & \textbf{multimix} 
& 9.87 
& 31.82 
& \cellcolor{purple!20}\textbf{8.64} 
& 58.55 
& \cellcolor{purple!20}\textbf{40.99} 
& \cellcolor{purple!20}\textbf{21.95} 
& 43.21 
& 38.88 
& 24.19 
& 30.90 \\

\textbf{Any4} & \textbf{multi} 
& 9.88 
& 31.91 
& 8.68 
& 58.91 
& \cellcolor{yellow!20}\underline{41.01} 
& 22.17 
& 43.67 
& 38.90 
& 24.38 
& 31.06 \\

\textbf{Any4} & \textbf{multi10} 
& 9.88 
& 31.78 
& 8.67 
& 58.14 
& 41.21 
& 22.06 
& \cellcolor{purple!20}\textbf{43.14} 
& \cellcolor{yellow!20}\underline{38.66} 
& \cellcolor{purple!20}\textbf{24.12} 
& \cellcolor{yellow!20}\underline{30.85} \\
\bottomrule
\end{tabular}
}
\caption{Perplexity results for Any4 4-bit quantization on Qwen2.5 across calibration strategies.
Multilingual and multilingual-mixture calibration yield consistently lower perplexity across
languages compared to monolingual calibration.}
\label{tab:any4_results}
\end{table*}

\begin{table}[htbp!]
\centering
\small
\begin{tabular}{lcc}
\toprule
\textbf{Language} & \textbf{XStoryCloze} $\boldsymbol{\rho}$ & \textbf{Global MMLU} $\boldsymbol{\rho}$ \\
\midrule
English (en)   & $-0.21$ & $-0.70$ \\
Swahili (sw)   & $-0.46$ & $-0.29$ \\
Chinese (zh)   & $-0.14$ & $-0.20$ \\
\bottomrule
\end{tabular}
\caption{Spearman rank correlation ($\rho$) between perplexity (PPL) and downstream accuracy
across calibration strategies. Negative values indicate that lower perplexity corresponds to
higher task performance.}
\label{tab:ppl_downstream_corr}
\end{table}

\subsection{Complete Per-Language Results for Llama3.1 8B on Wikipedia}

Full Llama3.1 8B on Wikipedia results reinforce the main pattern: multilingual mixes win on averages, while language-matched AWQ often yields the best per-language minima. \Cref{tab:awq_full_table} lists AWQ runs (plus FP16 and uniform 4-bit). Multilingual \texttt{multi}/\texttt{multi10} are consistently strong on the Avg column; language-matched (e.g., \texttt{xh}) often edges the best per-language scores; and code/math-enriched multilingual sets are competitive or best on average—supporting the "rare tokens help" claim.

For GPTQ, broad coverage matters most, and code/math boosts help. In \Cref{tab:gptq_full_table}, \texttt{multi10} is the best plain mix, and \texttt{codemath-multi10} often yields the best overall average, underscoring GPTQ’s sensitivity to calibration diversity and to tokens that expand activation tails (digits, brackets, symbols).

\begin{table*}[htbp!]
\centering
\resizebox{\textwidth}{!}{
\begin{tabular}{l|l|rrrrrrrrrrr}
\textbf{Quantization} & \textbf{Calibration} &
  \multicolumn{1}{l}{\textbf{en}} &
  \multicolumn{1}{l}{\textbf{sw}} &
  \multicolumn{1}{l}{\textbf{fr}} &
  \multicolumn{1}{l}{\textbf{xh}} &
  \multicolumn{1}{l}{\textbf{zh}} &
  \multicolumn{1}{l}{\textbf{st}} &
  \multicolumn{1}{l}{\textbf{yo}} &
  \multicolumn{1}{l}{\textbf{zu}} &
  \multicolumn{1}{l}{\textbf{ha}} &
  \multicolumn{1}{l}{\textbf{ig}} &
  \multicolumn{1}{l}{\textbf{Avg}} \\ \toprule

\textbf{FP16} & -- & 7.327 & 6.510 & 5.698 & 9.526 & 69.300 & 17.579 & 10.148 & 13.410 & 11.013 & 7.645 & 15.816 \\
\textbf{Uniform INT4} & -- & 8.327 & 7.836 & 6.323 & 11.085 & 84.210 & 23.266 & 12.551 & 17.053 & 15.670 & 10.151 & 19.647 \\ \hline

% ==================== AWQ: Natural (1 bold + 1 underline per column) ====================
\textbf{AWQ} & \textbf{en} 
& 7.695 & 5.936 & 5.939 & 9.989 & 66.536 & 15.941 & 8.822 & 10.198 & 9.915 & 7.816 & 14.879 \\
\textbf{AWQ} & \textbf{sw} 
& \underline{7.679} & \cellcolor{purple!20}\textbf{5.859} & 5.944 & 9.983 & 65.762 & 15.702 & 8.804 & 10.103 & 9.859 & 7.695 & 14.739 \\
\textbf{AWQ} & \textbf{fr} 
& 7.693 & 5.950 & \textbf{5.913} & 9.975 & 67.328 & 16.127 & 8.841 & 10.365 & 9.997 & 7.855 & 15.004 \\
\textbf{AWQ} & \textbf{xh} 
& 7.703 & 5.914 & 5.951 & 10.005 & \cellcolor{purple!20}\textbf{64.254} & \textbf{15.524} & \cellcolor{yellow!20}\underline{8.721} & \cellcolor{purple!20}\textbf{9.811} & 9.815 & \cellcolor{purple!20}\textbf{7.630} & \cellcolor{purple!20}\textbf{14.533} \\
\textbf{AWQ} & \textbf{zh} 
& 7.683 & 5.924 & 5.938 & \underline{9.888} & 66.004 & 15.901 & 8.887 & 10.126 & 9.982 & 7.772 & 14.810 \\
\textbf{AWQ} & \textbf{multi} 
& \textbf{7.675} & 5.894 & \textbf{5.913} & 9.932 & \underline{64.981} & 15.843 & 8.757 & \underline{9.962} & \cellcolor{yellow!20}\textbf{9.753} & \underline{7.676} & \underline{14.639} \\
\textbf{AWQ} & \textbf{multi10} 
& 7.688 & \underline{5.874} & \underline{5.926} & 9.994 & 65.773 & \underline{15.590} & \cellcolor{purple!20}\textbf{8.706} & 10.040 & \underline{9.788} & 7.695 & 14.707 \\
\textbf{AWQ} & \textbf{multimix} 
& 7.683 & 5.913 & 5.929 & \textbf{9.970} & 65.662 & 15.702 & 8.789 & 10.050 & 9.851 & 7.724 & 14.727 \\ \hline

% ==================== AWQ: Code (1 bold + 1 underline per column) ====================
\textbf{AWQ} & \textbf{code} 
& 7.666 & 5.926 & 5.935 & 9.959 & 66.480 & 15.737 & 8.744 & 10.171 & 9.925 & 7.735 & 14.828 \\
\textbf{AWQ} & \textbf{code-en} 
& \cellcolor{yellow!20}\underline{7.658} & 5.927 & 5.925 & 9.949 & 65.907 & 15.715 & 8.768 & 10.034 & 9.878 & 7.776 & 14.754 \\
\textbf{AWQ} & \textbf{code-sw} 
& 7.676 & \underline{5.876} & 5.945 & 9.988 & 66.097 & 15.959 & 8.814 & 10.171 & 9.928 & \underline{7.689} & 14.814 \\
\textbf{AWQ} & \textbf{code-fr} 
& 7.675 & 5.924 & \cellcolor{purple!20}\textbf{5.905} & 9.959 & 66.631 & 15.933 & 8.820 & 10.255 & 9.858 & 7.730 & 14.869 \\
\textbf{AWQ} & \textbf{code-xh} 
& 7.688 & 5.906 & 5.944 & 9.994 & \textbf{64.516} & \cellcolor{purple!20}\textbf{15.374} & \underline{8.743} & \textbf{9.846} & 9.802 & 7.758 & 14.557 \\
\textbf{AWQ} & \textbf{code-zh} 
& 7.677 & 5.927 & 5.928 & \cellcolor{yellow!20}\textbf{9.881} & 65.805 & 15.761 & 8.888 & 10.086 & 9.952 & 7.808 & 14.771 \\
\textbf{AWQ} & \textbf{code-multi} 
& 7.663 & 5.901 & 5.918 & \underline{9.932} & 65.792 & 15.675 & 8.770 & \underline{9.982} & \textbf{9.762} & \textbf{7.648} & \underline{14.704} \\
\textbf{AWQ} & \textbf{code-multi10} 
& \cellcolor{purple!20}\textbf{7.657} & \cellcolor{yellow!20}\textbf{5.864} & 5.924 & 9.966 & \underline{65.541} & \underline{15.482} & \textbf{8.742} & 10.020 & \underline{9.763} & 7.691 & \textbf{14.665} \\
\textbf{AWQ} & \textbf{code-multimix} 
& 7.667 & 5.928 & \underline{5.917} & 9.959 & 65.846 & 15.695 & 8.790 & 10.071 & 9.900 & 7.746 & 14.752 \\ \hline

% ==================== AWQ: Math (1 bold + 1 underline per column) ====================
\textbf{AWQ} & \textbf{math} 
& 7.743 & 6.013 & 5.982 & 10.091 & 66.868 & 16.055 & 8.892 & 10.287 & 10.077 & 7.887 & 14.990 \\
\textbf{AWQ} & \textbf{math-en} 
& 7.685 & 5.918 & 5.925 & 9.981 & 66.263 & 15.884 & 8.775 & 10.181 & 9.903 & 7.799 & 14.831 \\
\textbf{AWQ} & \textbf{math-sw} 
& 7.691 & \textbf{5.872} & 5.942 & 9.988 & 65.941 & 15.716 & \underline{8.747} & 10.142 & 9.863 & \cellcolor{yellow!20}\textbf{7.640} & 14.754 \\
\textbf{AWQ} & \textbf{math-fr} 
& \underline{7.675} & 5.935 & \cellcolor{yellow!20}\textbf{5.908} & 9.970 & 65.880 & 15.916 & 8.886 & 10.093 & 9.787 & 7.726 & 14.778 \\
\textbf{AWQ} & \textbf{math-xh} 
& 7.694 & 5.922 & 5.945 & 10.012 & \cellcolor{yellow!20}\textbf{64.327} & \underline{15.505} & 8.777 & \cellcolor{yellow!20}\textbf{9.818} & 9.800 & 7.719 & \textbf{14.552} \\
\textbf{AWQ} & \textbf{math-zh} 
& 7.687 & 5.930 & 5.937 & \textbf{9.886} & 66.490 & 15.865 & 8.886 & 10.182 & 9.956 & 7.737 & 14.856 \\
\textbf{AWQ} & \textbf{math-multi} 
& \textbf{7.674} & 5.889 & \underline{5.915} & \underline{9.934} & \underline{65.154} & 15.574 & \textbf{8.718} & \underline{9.951} & \cellcolor{purple!20}\textbf{9.746} & \underline{7.673} & \underline{14.623} \\
\textbf{AWQ} & \textbf{math-multi10} 
& 7.694 & \underline{5.883} & 5.937 & 9.986 & 65.614 & \textbf{15.441} & 8.762 & 10.035 & 9.815 & 7.689 & 14.686 \\
\textbf{AWQ} & \textbf{math-multimix} 
& 7.690 & 5.928 & 5.934 & 9.972 & 65.834 & 15.758 & 8.812 & 10.100 & \underline{9.767} & 7.702 & 14.750 \\ \hline

% ==================== AWQ: Codemath (1 bold + 1 underline per column) ====================
\textbf{AWQ} & \textbf{codemath} 
& \underline{7.672} & 5.934 & 5.936 & 9.964 & 67.009 & 15.994 & 8.785 & 10.197 & 9.901 & 7.759 & 14.915 \\
\textbf{AWQ} & \textbf{codemath-en} 
& 7.681 & 5.917 & 5.927 & 9.968 & 66.284 & 15.805 & \underline{8.731} & 10.225 & 9.944 & 7.818 & 14.830 \\
\textbf{AWQ} & \textbf{codemath-sw} 
& 7.694 & \textbf{5.873} & 5.947 & 9.991 & 66.149 & 15.849 & 8.758 & 10.140 & 9.864 & \underline{7.686} & 14.795 \\
\textbf{AWQ} & \textbf{codemath-fr} 
& \underline{7.672} & 5.934 & \textbf{5.909} & 9.976 & 66.358 & 15.906 & 8.823 & 10.165 & 9.871 & 7.747 & 14.836 \\
\textbf{AWQ} & \textbf{codemath-xh} 
& 7.687 & 5.908 & 5.946 & 9.998 & \textbf{64.352} & \cellcolor{yellow!20}\textbf{15.405} & 8.751 & \textbf{9.827} & \underline{9.800} & 7.704 & \cellcolor{yellow!20}\textbf{14.538} \\
\textbf{AWQ} & \textbf{codemath-zh} 
& 7.684 & 5.937 & 5.935 & \cellcolor{purple!20}\textbf{9.879} & 65.811 & 15.685 & 8.830 & 10.090 & 9.900 & 7.737 & 14.749 \\
\textbf{AWQ} & \textbf{codemath-multi} 
& \textbf{7.664} & \underline{5.896} & 5.921 & \underline{9.937} & \underline{65.460} & \underline{15.499} & \textbf{8.730} & \underline{10.001} & 9.811 & \textbf{7.683} & \underline{14.660} \\
\textbf{AWQ} & \textbf{codemath-multi10} 
& 7.675 & \textbf{5.873} & \underline{5.918} & 9.975 & 65.944 & 15.679 & 8.745 & 10.048 & \textbf{9.790} & 7.711 & 14.736 \\
\textbf{AWQ} & \textbf{codemath-multimix} 
& 7.686 & 5.915 & 5.933 & 9.966 & 66.047 & 15.573 & 8.853 & 10.091 & 9.808 & 7.716 & 14.759 \\
\bottomrule
\end{tabular}
}
\caption{Llama3.1 8B Per-language perplexity for AWQ (plus FP16 and uniform INT4 baselines) on the Wikipedia evaluation set (lower is better). 
Within each block (natural / code / math / codemath), \textbf{bold} marks the best and \underline{underline} marks the second-best per column. Across all AWQ rows, \textcolor{black}{\fcolorbox{white}{purple!20}{\strut}} marks the best overall per column and \textcolor{black}{\fcolorbox{white}{yellow!20}{\strut}} marks the second-best overall; global-best cells use \textbf{bold+purple}. Multilingual calibration (\texttt{multi}/\texttt{multi10}) achieves top or near-top averages; language-matched AWQ often wins per-language minima; and code/math additions to multilingual sets provide small extra gains on average.}
\label{tab:awq_full_table}
\end{table*}

\begin{table*}[htbp!]
\centering
\resizebox{\textwidth}{!}{
\begin{tabular}{l|l|rrrrrrrrrrr}
\textbf{Quantization} & \textbf{Calibration} &
  \multicolumn{1}{l}{\textbf{en}} &
  \multicolumn{1}{l}{\textbf{sw}} &
  \multicolumn{1}{l}{\textbf{fr}} &
  \multicolumn{1}{l}{\textbf{xh}} &
  \multicolumn{1}{l}{\textbf{zh}} &
  \multicolumn{1}{l}{\textbf{st}} &
  \multicolumn{1}{l}{\textbf{yo}} &
  \multicolumn{1}{l}{\textbf{zu}} &
  \multicolumn{1}{l}{\textbf{ha}} &
  \multicolumn{1}{l}{\textbf{ig}} &
  \multicolumn{1}{l}{\textbf{Avg}} \\ \toprule
\textbf{FP16} & -- & 7.327 & 6.510 & 5.698 & 9.526 & 69.300 & 17.579 & 10.148 & 13.410 & 11.013 & 7.645 & 15.816 \\
\textbf{Uniform INT4} & -- & 8.327 & 7.836 & 6.323 & 11.085 & 84.210 & 23.266 & 12.551 & 17.053 & 15.670 & 10.151 & 19.647 \\ \hline

% ==================== GPTQ: Natural ====================
\textbf{GPTQ} & \textbf{en} & 8.3 & 8.73 & 6.685 & 12.242 & 89.283 & 28.453 & 13.022 & 17.829 & 18.999 & 12.696 & 21.624 \\
\textbf{GPTQ} & \textbf{sw} & 9.286 & 9.633 & 7.908 & 15.054 & 95.637 & 35.093 & 15.035 & 19.274 & 23.086 & 18.786 & 24.879 \\
\textbf{GPTQ} & \textbf{fr} & 8.329 & 8.519 & 6.41 & 12.49 & 86.817 & 28.205 & 12.907 & 17.306 & 18.068 & 12.756 & 21.181 \\
\textbf{GPTQ} & \textbf{xh} & 8.612 & 7.844 & 6.693 & 12.538 & \underline{79.847} & 25.564 & 12.671 & 15.714 & 17.028 & 11.99 & 19.85 \\
\textbf{GPTQ} & \textbf{zh} & 8.489 & 8.523 & 6.596 & 11.228 & 89.869 & 25.902 & 12.816 & 18.194 & 17.925 & 13.854 & 21.34 \\
\textbf{GPTQ} & \textbf{multimix} & \underline{8.189} & \underline{7.87} & 6.291 & 11.22 & 81.481 & \underline{24.05} & 12.294 & \underline{16.27} & 15.664 & 11.359 & 19.469 \\
\textbf{GPTQ} & \textbf{multi} & 8.211 & 7.604 & \underline{6.26} & \underline{11.05} & 82.564 & 24.725 & \underline{11.89} & 16.304 & \underline{14.545} & \underline{10.06} & \underline{19.321} \\
\textbf{GPTQ} & \textbf{multi10} & \textbf{8.184} & \textbf{7.222} & \textbf{6.194} & \textbf{10.787} & \textbf{77.444} & \textbf{20.693} & \cellcolor{yellow!20}\textbf{11.329} & \textbf{15.176} & \textbf{14.296} & \textbf{9.714} & \textbf{18.104} \\

\hline

% ==================== GPTQ: Code ====================
\textbf{GPTQ} & \textbf{code} & 8.268 & 7.768 & 6.364 & 11.089 & 82.555 & \textbf{22.846} & 12.059 & 16.544 & \underline{14.834} & \underline{10.293} & \underline{19.262} \\
\textbf{GPTQ} & \textbf{code-en} & 8.302 & 8.476 & 6.656 & 11.806 & 88.28 & 28.43 & 12.585 & 17.729 & 18.043 & 13.355 & 21.366 \\
\textbf{GPTQ} & \textbf{code-sw} & 8.436 & \textbf{7.45} & 6.583 & 11.738 & 81.903 & 26.098 & 12.403 & 16.265 & 16.904 & 11.466 & 19.925 \\
\textbf{GPTQ} & \textbf{code-fr} & 8.238 & 7.88 & \underline{6.245} & 11.32 & 82.009 & 24.961 & 12.126 & 16.246 & 15.536 & 11.112 & 19.567 \\
\textbf{GPTQ} & \textbf{code-xh} & 8.472 & 7.967 & 6.617 & 12.206 & \underline{78.896} & 25.842 & 12.316 & \textbf{15.621} & 16.928 & 12.123 & 19.699 \\
\textbf{GPTQ} & \textbf{code-zh} & 8.292 & 8.161 & 6.459 & \underline{11.061} & 84.575 & 25.325 & 12.224 & 16.834 & 16.647 & 11.946 & 20.152 \\
\textbf{GPTQ} & \textbf{code-multi} & \textbf{8.191} & \underline{7.496} & \textbf{6.214} & \textbf{10.95} & \textbf{79.968} & \underline{22.969} & \textbf{11.547} & \underline{15.836} & \textbf{13.895} & \textbf{9.699} & \textbf{18.676} \\
\textbf{GPTQ} & \textbf{code-multi10} & 8.462 & 7.971 & 6.515 & 11.742 & 83.681 & 26.239 & \underline{11.951} & 16.459 & 17.113 & 11.695 & 20.183 \\
\textbf{GPTQ} & \textbf{code-multimix} & \underline{8.286} & 8.289 & 6.393 & 11.498 & 84.651 & 25.529 & 12.32 & 16.907 & 17.342 & 11.769 & 20.298 \\

\hline

% ==================== GPTQ: Math ====================
\textbf{GPTQ} & \textbf{math} & 9.064 & 10.196 & 7.378 & 13.528 & 100.6 & 32.712 & 14.731 & 20.347 & 24.035 & 16.717 & 24.931 \\
\textbf{GPTQ} & \textbf{math-en} & 8.263 & 9.037 & 6.593 & 11.809 & 90.716 & 29.414 & 12.837 & 18.077 & 19.416 & 12.431 & 21.859 \\
\textbf{GPTQ} & \textbf{math-sw} & 8.813 & 8.45 & 7.188 & 13.721 & 90.164 & 30.551 & 13.625 & 18.066 & 20.476 & 14.857 & 22.591 \\
\textbf{GPTQ} & \textbf{math-fr} & 8.216 & 7.882 & \underline{6.193} & 11.356 & \underline{80.508} & 24.129 & 12.167 & 16.069 & 15.69 & 11.156 & 19.337 \\
\textbf{GPTQ} & \textbf{math-xh} & 8.334 & \textbf{7.295} & 6.381 & 11.295 & \cellcolor{purple!20}\textbf{73.26} & \cellcolor{purple!20}\textbf{20.281} & \textbf{11.52} & \cellcolor{purple!20}\textbf{14.515} & \textbf{13.904} & \cellcolor{yellow!20}\textbf{9.528} & \cellcolor{purple!20}\textbf{17.631} \\
\textbf{GPTQ} & \textbf{math-zh} & 8.338 & 8.28 & 6.471 & \underline{10.894} & 87.605 & 27.126 & 12.385 & 17.504 & 17.501 & 12.747 & 20.885 \\
\textbf{GPTQ} & \textbf{math-multi10} & 8.398 & 7.805 & 6.458 & 11.575 & 82.704 & 24.904 & 11.855 & 16.362 & 17.17 & 11.6 & 19.883 \\
\textbf{GPTQ} & \textbf{math-multi} & \underline{8.19} & \underline{7.523} & \cellcolor{yellow!20}\textbf{6.188} & \textbf{10.851} & 80.772 & \underline{23.11} & \underline{11.6} & \underline{16.039} & \underline{14.069} & \underline{9.846} & \underline{18.819} \\
\textbf{GPTQ} & \textbf{math-multimix} & \textbf{8.178} & 7.963 & 6.303 & 11.379 & 81.328 & 24.029 & 12.102 & 16.25 & 16.273 & 10.924 & 19.473 \\

\hline

% ==================== GPTQ: Codemath ====================
\textbf{GPTQ} & \textbf{codemath} & 8.349 & 8.225 & 6.438 & 11.217 & 84.687 & 25.21 & 12.347 & 17.313 & 16.325 & 11.408 & 20.152 \\
\textbf{GPTQ} & \textbf{codemath-en} & 8.199 & 8.134 & 6.414 & 11.424 & 83.765 & 25.346 & 12.374 & 16.824 & 17.464 & 11.452 & 20.14 \\
\textbf{GPTQ} & \textbf{codemath-sw} & 8.418 & 7.557 & 6.643 & 12.108 & 82.33 & 26.268 & 12.616 & 16.257 & 16.798 & 12.135 & 20.113 \\
\textbf{GPTQ} & \textbf{codemath-fr} & 8.341 & 8.367 & 6.334 & 11.606 & 86.197 & 27.064 & 12.713 & 17.342 & 17.631 & 12.189 & 20.778 \\
\textbf{GPTQ} & \textbf{codemath-xh} & 8.348 & 7.571 & 6.43 & 11.375 & \cellcolor{yellow!20}\textbf{76.623} & 23.131 & 12.08 & \underline{15.219} & 15.316 & 11.045 & 18.714 \\
\textbf{GPTQ} & \textbf{codemath-zh} & 8.287 & 8.013 & 6.38 & \cellcolor{purple!20}\textbf{10.677} & 83.715 & 24.378 & 12.067 & 16.521 & 15.908 & 10.579 & 19.652 \\
\textbf{GPTQ} & \textbf{codemath-multi} & 8.174 & \underline{7.495} & \underline{6.193} & 10.744 & 80.949 & 22.026 & \underline{11.61} & 16.203 & \cellcolor{yellow!20}\underline{13.838} & \underline{9.534} & 18.677 \\
\textbf{GPTQ} & \textbf{codemath-multi10} & \cellcolor{yellow!20}\underline{8.159} & \cellcolor{purple!20}\textbf{7.176} & \cellcolor{purple!20}\textbf{6.183} & \cellcolor{yellow!20}\underline{10.695} & \underline{76.654} & \cellcolor{yellow!20}\textbf{20.648} & \cellcolor{purple!20}\textbf{11.237} & \cellcolor{yellow!20}\textbf{15.083} & \cellcolor{purple!20}\textbf{13.803} & \cellcolor{purple!20}\textbf{9.464} & \cellcolor{yellow!20}\textbf{17.91} \\
\textbf{GPTQ} & \textbf{codemath-multimix} & \cellcolor{purple!20}\textbf{8.141} & 7.672 & 6.28 & 11.052 & 78.737 & \underline{21.988} & 11.715 & 15.668 & 14.431 & 10.089 & \underline{18.577} \\

\bottomrule
\end{tabular}
}
\caption{Llama3.1 8B per-language perplexity for GPTQ (plus FP16 and uniform INT4 baselines) on the Wikipedia evaluation set (lower is better). Within each block (Natural / Code / Math / Codemath), \textbf{bold} marks the best and \underline{underline} marks the second-best per column. Across all GPTQ rows (all blocks combined), \textcolor{black}{\fcolorbox{white}{purple!20}{\strut}} marks the best overall per column and \textcolor{black}{\fcolorbox{white}{yellow!20}{\strut}} marks the second-best overall. \texttt{multi10} is the strongest plain multilingual mix; code/math-enriched multilingual sets (e.g., \texttt{codemath-multi10}) deliver the best averages overall, highlighting GPTQ's gain from diverse, outlier-rich calibration.}
\label{tab:gptq_full_table}
\end{table*}

\subsection{Cross-Lingual Transfer}

Our results provide evidence of language-family transfer as we cover languages within the same family: English, French (Indo-European), Swahili, isiXhosa, Sotho, Zulu(Bantu families). As shown in \Cref{tab:calibration_pplx_combined}, when calibrating using isiXhosa (xh), we observe improvements not only for isiXhosa itself, but also for structurally related Bantu languages (Zulu and Sotho), compared to English-calibrated baselines. We also see the same within-family transfer pattern under GPTQ.

\begin{table*}[htbp!]
\centering
\resizebox{\textwidth}{!}{
\begin{tabular}{lcccccc}
\toprule
\textbf{Calibration Language} 
& \multicolumn{3}{c}{\textbf{AWQ }} 
& \multicolumn{3}{c}{\textbf{GPTQ }} \\
\cmidrule(lr){2-4} \cmidrule(lr){5-7}
& \textbf{xh} & \textbf{en} & \textbf{Improvement ($\downarrow$)}
& \textbf{xh} & \textbf{en} & \textbf{Improvement ($\downarrow$)} \\
\midrule
isiXhosa (xh) & 64.254 & 66.536 & 2.282 & 79.847 & 89.283 & 9.436 \\
Zulu (zu)     & 9.811  & 10.198 & 0.387 & 15.714 & 17.829 & 2.115 \\
Sotho (st)    & 15.524 & 15.941 & 0.417 & 25.564 & 28.453 & 2.889 \\
\bottomrule
\end{tabular}}
\caption{Perplexity comparison of calibration languages for AWQ and GPTQ across evaluation languages for Llama3.1 8B. Lower values indicate better performance.}
\label{tab:calibration_pplx_combined}
\end{table*}

\subsection{Perplexity--Downstream Correlation}
\label{sec:perp_downcorrelation}

To directly quantify the relationship between language modeling quality and task-level behavior,
we analyze the rank correlation between perplexity (PPL) and downstream accuracy using the
XStoryCloze results reported in the main paper and the Global MMLU results introduced in this
revision.
For each language where both metrics are available, we compute Spearman’s rank correlation
coefficient across calibration strategies.
As shown in Table~\ref{tab:ppl_downstream_corr}, we observe a consistent negative correlation
across languages and tasks, indicating that lower perplexity reliably corresponds to higher
downstream accuracy.
This confirms that calibration-induced PPL improvements translate into functional task gains,
rather than reflecting purely distributional artifacts.

\subsection{Translated vs.\ Native Calibration Data}
\label{sec:appendix_translation}

Several single-language calibration sets in our experiments are constructed by machine-translating the English calibration corpus. This design enables a controlled isolation of language effects while holding lexical content and calibration budget fixed. To ensure that our conclusions are not driven by artifacts of machine translation, we additionally perform calibration using \emph{native} multilingual corpora—specifically, language-specific subsets of C4 and Wikipedia—where calibration data is drawn from naturally occurring text. Corresponding results are reported in \Cref{tab:wiki_llama_controlled,tab:wiki_qwen_controlled,tab:c4_llama_controlled,tab:c4_qwen_controlled}. Across both Llama3.1 8B and Qwen2.5 7B, and for both GPTQ and AWQ, we observe consistent qualitative behavior between translated and native calibration sources. In particular, language-diverse calibration consistently outperforms English-only baselines, and within-family transfer effects (e.g., among Bantu languages) remain stable. These findings indicate that the benefits of language-specific calibration reflect genuine linguistic and distributional properties of the
calibration data, rather than effects specific to machine-translated text.

\subsection{Controlled Calibration Design: Same-Corpus Calibration for GPTQ and AWQ}
\label{sec:appendix_controlled_calibration}

Our primary calibration choices—C4 for GPTQ and the Pile validation split for AWQ—follow the default preprocessing pipelines and recommended usage in the respective method papers and reference implementations. To verify that the behavioral differences observed between GPTQ and AWQ are not artifacts of calibrating on different corpora, we conduct controlled experiments in which \emph{both} quantizers are calibrated on the \emph{same} underlying corpus, while varying only the calibration language composition. Concretely, we evaluate (i) Wikipedia$\rightarrow$Wikipedia calibration and (ii) C4$\rightarrow$C4 calibration for both GPTQ and AWQ, across two model families (Llama3.1 8B and Qwen2.5 7B). Results in \Cref{tab:wiki_llama_controlled,tab:wiki_qwen_controlled,tab:c4_llama_controlled,tab:c4_qwen_controlled} show that the qualitative differences between GPTQ and AWQ persist under identical calibration corpora: multilingual and multilingual-mixture calibration improves global robustness, while language-matched calibration yields localized gains. This confirms that the quantizer-specific effects reported in the main text arise from genuine algorithmic differences rather than corpus selection.

\begin{table*}[htbp]
\centering
\small
\resizebox{\textwidth}{!}{%
\begin{tabular}{llccccccccccc}
\toprule
\textbf{Quantizer} & \textbf{Calibration} &
\textbf{en} & \textbf{sw} & \textbf{fr} & \textbf{zh} & \textbf{st} & \textbf{xh} & \textbf{yo} & \textbf{zu} & \textbf{ha} & \textbf{ig} & \textbf{Avg} \\
\midrule
AWQ  & en & 7.641 & 5.911 & 5.921 & 9.961 & 15.823 & 65.947 & 8.772 & 10.109 & 9.850 & 7.756 & 14.769 \\
AWQ  & fr & 7.671 & 5.916 & 5.905 & 9.974 & 15.666 & 65.653 & 8.762 & 10.061 & 9.838 & 7.692 & 14.714 \\
AWQ  & sw & 7.691 & 5.864 & 5.944 & 9.978 & 15.699 & 65.921 & 8.728 & 10.121 & 9.946 & 7.672 & 14.756 \\
AWQ  & xh & 7.707 & 5.913 & 5.948 & 10.017 & 15.423 & 63.682 & 8.729 & 9.768 & 9.848 & 7.666 & 14.470 \\
AWQ  & zh & 7.691 & 5.917 & 5.936 & 9.861 & 15.851 & 66.348 & 8.818 & 10.187 & 9.978 & 7.811 & 14.840 \\
\midrule
GPTQ & en & 7.994 & 7.658 & 6.286 & 11.623 & 23.507 & 82.570 & 11.884 & 16.187 & 14.979 & 10.876 & 19.356 \\
GPTQ & sw & 9.131 & 9.905 & 7.444 & 15.010 & 39.161 & 98.595 & 14.577 & 19.890 & 25.060 & 19.789 & 25.856 \\
GPTQ & fr & 8.109 & 7.708 & 6.137 & 11.107 & 23.952 & 82.231 & 11.896 & 16.373 & 14.901 & 10.726 & 19.314 \\
GPTQ & xh & 8.364 & 7.692 & 6.519 & 11.855 & 25.236 & 78.434 & 12.009 & 15.468 & 16.669 & 11.515 & 19.376 \\
GPTQ & zh & 8.510 & 8.516 & 6.671 & 11.383 & 27.828 & 87.892 & 12.454 & 17.846 & 17.557 & 13.385 & 21.204 \\
\bottomrule
\end{tabular}
}
\caption{Wikipedia$\rightarrow$Wikipedia controlled calibration results (perplexity) for Llama3.1 8B, where both GPTQ and AWQ are calibrated on the same corpus.}
\label{tab:wiki_llama_controlled}
\end{table*}

\begin{table*}[htbp]
\centering
\small
\resizebox{\textwidth}{!}{%
\begin{tabular}{llccccccccccc}
\toprule
\textbf{Quantizer} & \textbf{Calibration} &
\textbf{en} & \textbf{sw} & \textbf{fr} & \textbf{zh} & \textbf{st} & \textbf{xh} & \textbf{yo} & \textbf{zu} & \textbf{ha} & \textbf{ig} & \textbf{Avg} \\
\midrule
AWQ  & en & 7.194 & 13.923 & 6.435 & 10.639 & 33.023 & 20.639 & 12.744 & 17.078 & 25.076 & 20.531 & 16.728 \\
AWQ  & fr & 7.212 & 13.842 & 6.350 & 10.568 & 32.997 & 20.599 & 12.637 & 17.088 & 25.213 & 20.448 & 16.695 \\
AWQ  & sw & 7.241 & 13.675 & 6.440 & 10.615 & 33.807 & 20.820 & 12.812 & 17.245 & 25.006 & 20.856 & 16.852 \\
AWQ  & xh & 7.272 & 13.727 & 6.452 & 10.713 & 33.737 & 20.619 & 12.824 & 17.036 & 25.200 & 20.852 & 16.843 \\
AWQ  & zh & 7.223 & 13.941 & 6.422 & 10.349 & 32.936 & 20.716 & 12.748 & 17.234 & 25.216 & 20.497 & 16.728 \\
\midrule
GPTQ & en & 13.440 & 26.310 & 8.108 & 13.693 & 51.723 & 50.512 & 25.755 & 29.092 & 28.213 & 16.910 & 26.376 \\
GPTQ & sw & 13.718 & 23.996 & 8.083 & 14.251 & 47.954 & 47.362 & 24.417 & 27.794 & 26.809 & 16.675 & 25.106 \\
GPTQ & fr & 13.644 & 26.118 & 7.877 & 14.253 & 50.867 & 49.204 & 24.704 & 28.817 & 28.021 & 16.949 & 26.046 \\
GPTQ & xh & 13.748 & 24.457 & 8.112 & 13.159 & 46.910 & 46.787 & 24.149 & 27.339 & 26.919 & 16.383 & 24.796 \\
GPTQ & zh & 13.937 & 26.027 & 8.271 & 12.704 & 50.932 & 49.071 & 25.080 & 28.879 & 27.735 & 16.661 & 25.930 \\
\bottomrule
\end{tabular}
}
\caption{Wikipedia$\rightarrow$Wikipedia controlled calibration results (perplexity) for Qwen2.5~7B, where both GPTQ and AWQ are calibrated on the same corpus.}
\label{tab:wiki_qwen_controlled}
\end{table*}

\begin{table*}[htbp]
\centering
\small
\resizebox{\textwidth}{!}{%
\begin{tabular}{llccccccccccc}
\toprule
\textbf{Quantizer} & \textbf{Calibration} &
\textbf{en} & \textbf{sw} & \textbf{fr} & \textbf{zh} & \textbf{xh} & \textbf{st} & \textbf{yo} & \textbf{zu} & \textbf{ha} & \textbf{ig} & \textbf{Avg} \\
\midrule
AWQ  & en & 12.496 & 11.695 & 8.893 & 13.072 & -- & 24.153 & 18.451 & 22.376 & 15.726 & 10.723 & 15.287 \\
AWQ  & fr & 12.471 & -- & 8.830 & 13.016 & 31.623 & 24.044 & 18.307 & 22.053 & 15.600 & 10.683 & 17.403 \\
AWQ  & sw & 12.552 & 11.449 & 8.892 & 13.050 & 31.623 & 23.619 & 18.200 & -- & 15.649 & 10.680 & 16.190 \\
AWQ  & xh & 12.514 & 11.575 & 8.885 & 12.966 & 30.999 & 23.517 & 17.946 & 21.697 & 15.531 & 10.580 & 16.621 \\
AWQ  & zh & 12.517 & 11.660 & 8.892 & 12.894 & 31.712 & 23.906 & 18.498 & 22.267 & 15.889 & 10.718 & 16.895 \\
\midrule
GPTQ & en & 15.473 & 16.319 & 10.767 & 17.861 & 43.863 & 40.179 & 23.115 & 32.704 & 25.826 & 14.551 & 24.066 \\
GPTQ & sw & 17.573 & 15.267 & 11.428 & 19.616 & 43.076 & 41.675 & 23.090 & 33.459 & 26.192 & 15.428 & 24.680 \\
GPTQ & fr & 15.289 & 14.028 & 9.843 & 16.236 & 39.774 & 32.883 & 21.540 & 30.626 & 22.193 & 13.211 & 21.562 \\
GPTQ & xh & 16.066 & 14.154 & 10.492 & 16.645 & 38.736 & 35.667 & 21.620 & 30.256 & 22.644 & 14.102 & 22.038 \\
GPTQ & zh & 16.038 & 14.845 & 10.488 & 15.082 & 40.798 & 33.604 & 20.983 & 32.145 & 23.598 & 13.642 & 22.122 \\
\bottomrule
\end{tabular}
}
\caption{C4$\rightarrow$C4 controlled calibration results (perplexity) for Llama3.1 8B. Missing entries indicate unavailable measurements in the provided runs.}
\label{tab:c4_llama_controlled}
\end{table*}

\begin{table*}[htbp]
\centering
\small
\resizebox{\textwidth}{!}{%
\begin{tabular}{llccccccccccc}
\toprule
\textbf{Quantizer} & \textbf{Calibration} &
\textbf{en} & \textbf{sw} & \textbf{fr} & \textbf{zh} & \textbf{xh} & \textbf{st} & \textbf{yo} & \textbf{zu} & \textbf{ha} & \textbf{ig} & \textbf{Avg} \\
\midrule
AWQ  & en & 13.524 & 24.613 & 7.905 & 12.554 & 31.807 & 36.384 & 26.660 & 24.429 & 29.204 & 17.006 & 22.408 \\
AWQ  & fr & 13.502 & 24.341 & 7.801 & 12.594 & 31.568 & 36.417 & 26.515 & 24.291 & 29.009 & 17.083 & 22.312 \\
AWQ  & sw & 13.597 & 23.786 & 7.914 & 12.546 & 31.391 & 36.418 & 26.480 & 24.271 & 28.983 & 17.300 & 22.269 \\
AWQ  & xh & 13.577 & 23.965 & 7.888 & 12.553 & 30.989 & 35.982 & 26.772 & 24.008 & 29.070 & 17.108 & 22.191 \\
AWQ  & zh & 13.528 & 24.286 & 7.922 & 12.308 & 31.445 & 36.263 & 26.846 & 24.178 & 29.196 & 17.071 & 22.304 \\
\midrule
GPTQ & en & 13.440 & 26.310 & 8.108 & 13.693 & 51.723 & 50.512 & 25.755 & 29.092 & 28.213 & 16.910 & 26.376 \\
GPTQ & sw & 13.718 & 23.996 & 8.083 & 14.251 & 47.954 & 47.362 & 24.417 & 27.794 & 26.809 & 16.675 & 25.106 \\
GPTQ & fr & 13.644 & 26.118 & 7.877 & 14.253 & 50.867 & 49.204 & 24.704 & 28.817 & 28.021 & 16.949 & 26.046 \\
GPTQ & xh & 13.748 & 24.457 & 8.112 & 13.159 & 46.910 & 46.787 & 24.149 & 27.339 & 26.919 & 16.383 & 24.796 \\
GPTQ & zh & 13.937 & 26.027 & 8.271 & 12.704 & 50.932 & 49.071 & 25.080 & 28.879 & 27.735 & 16.661 & 25.930 \\
\bottomrule
\end{tabular}
}
\caption{C4$\rightarrow$C4 controlled calibration results (perplexity) for Qwen2.5~7B, where both GPTQ and AWQ are calibrated on the same corpus.}
\label{tab:c4_qwen_controlled}
\end{table*}

\subsection{Impact of Quantization on Layers}
\begin{figure*}[htbp!]
  \centering

  \begin{subfigure}[b]{0.45\textwidth}
    \includegraphics[width=0.95\linewidth]{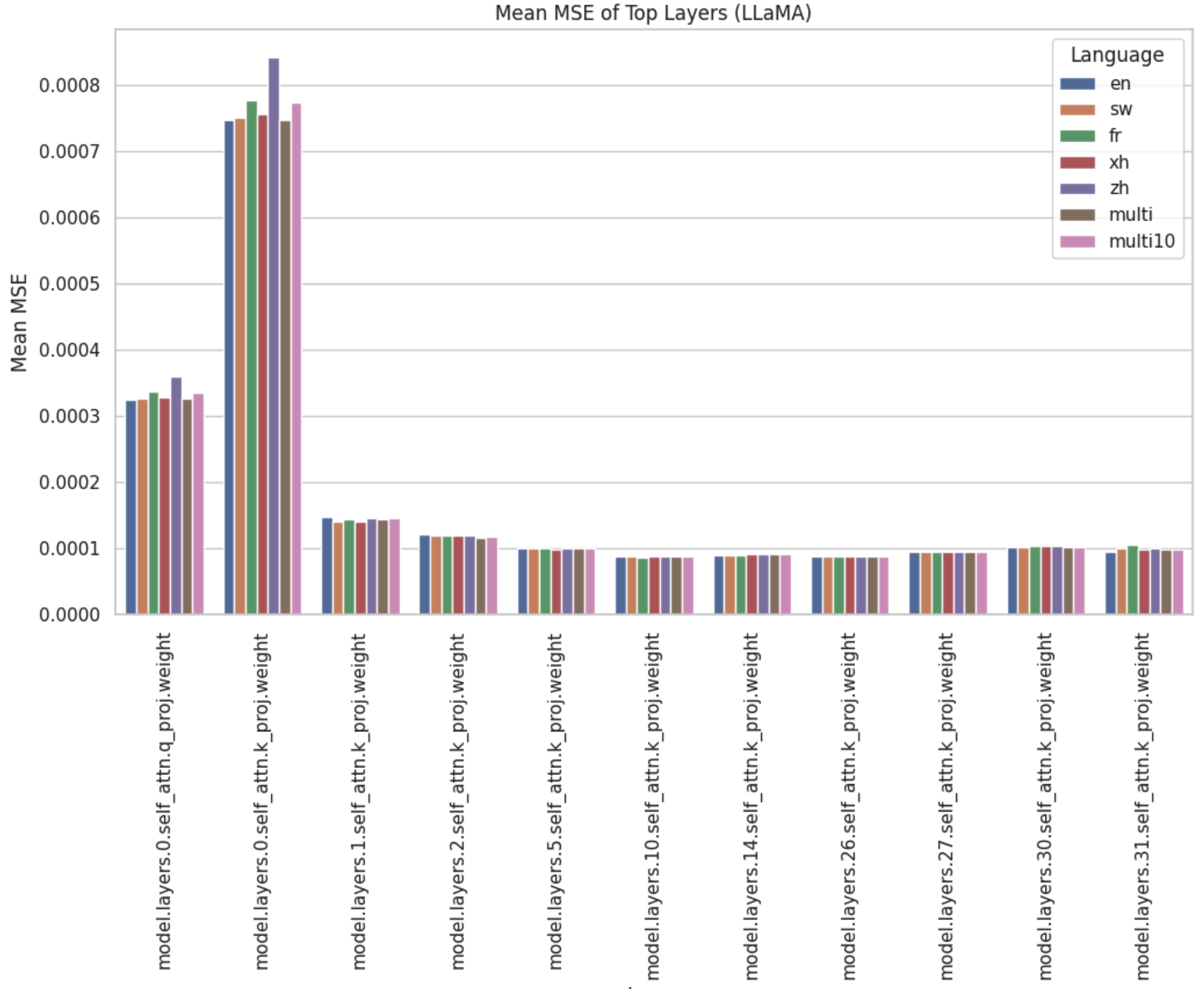}
    \caption{Llama3.1 8B}
    \label{fig:lay1}
  \end{subfigure}
  \begin{subfigure}[b]{0.45\textwidth}
    \includegraphics[width=0.87\linewidth]{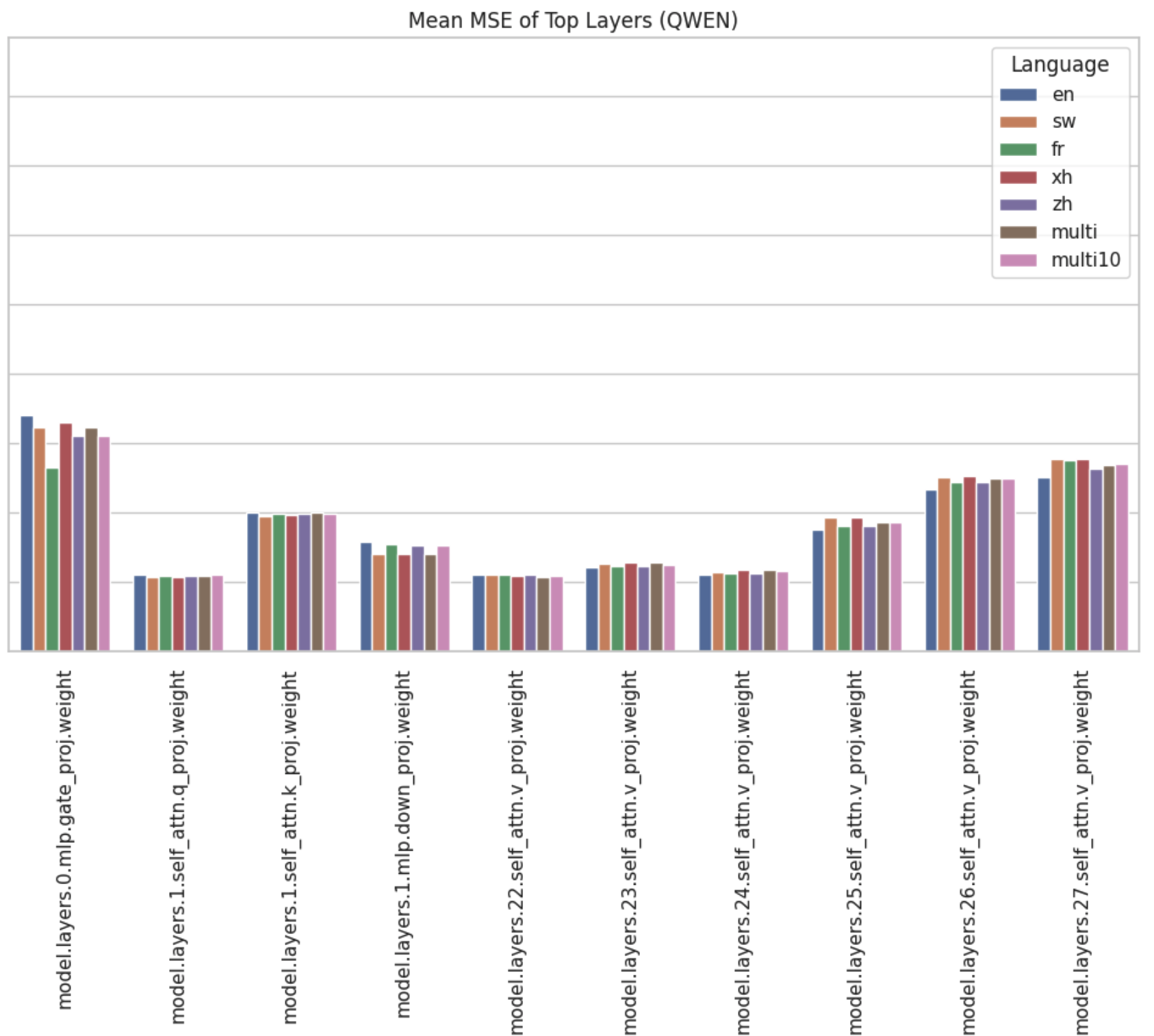}
    \caption{Qwen2.5 7B}
    \label{fig:lay2}
  \end{subfigure}
  \caption{Bars give the mean-squared error (MSE) between original and quantized weights for the most error-prone tensors in (a) Llama3.1 8B and (b) Qwen2.5 7B, colour-coded by calibration set (\texttt{en}, \texttt{sw}, \texttt{fr}, \texttt{xh}, \texttt{zh}, \texttt{multi}, \texttt{multi10}). Errors are more prominent in on layer 0 and for Qwen last layers.}
  \label{fig:layers}
\end{figure*}

\textit{Quantization error concentrates in the extremal layers, and its magnitude varies with the calibration set.} Beyond perplexity,we measure weight MSE and inspect activation statistics across layers to see where calibration matters most. For both models, the largest errors appear in the first and last transformer blocks—layers 0 and 31 for Llama3.1 8B and layers 0 and 27 for Qwen2.5 7B—providing a focused lens for algorithm–data interactions. In \Cref{fig:layers}, bars give the mean-squared error between original and quantized weights for the most error-prone tensors, color-coded by calibration set.

\end{document}